# Repeatability of Multiparametric Prostate MRI Radiomics Features

## Authors

Michael Schwier,[1,2] Joost van Griethuysen,[3] Mark G Vangel,[2,4] Steve Pieper,[5] Sharon Peled,[2,3] Clare M Tempany,[1,2] Hugo JWL Aerts,[2,6] Ron Kikinis,[1,2] Fiona M Fennessy,[1,2,6] and Andrey Fedorov[1,2]

**Contact**: mschwier@bwh.harvard.edu, andrey.fedorov@gmail.com

[1]Brigham and Women's Hospital, Boston, MA, United States
[2]Harvard Medical School, Boston, MA, United States
[3]Netherlands Cancer Institute Maastricht University, Amsterdam, Netherlands
[4]Massachusetts General Hospital, Charlestown, MA, United States
[5]Isomics, Inc., Cambridge, MA, United States
[6]Dana-Farber Cancer Institute, Boston, MA, United States

## Abstract

In this study we assessed the repeatability of the values of radiomics features for small prostate tumors using test-retest Multiparametric Magnetic Resonance Imaging (mpMRI) images. The premise of radiomics is that quantitative image features can serve as biomarkers characterizing disease. For such biomarkers to be useful, repeatability is a basic requirement, meaning its value must remain stable between two scans, if the conditions remain stable. We investigated repeatability of radiomics features under various preprocessing and extraction configurations including various image normalization schemes, different image pre-filtering, 2D vs 3D texture computation, and different bin widths for image discretization. Image registration as means to re-identify regions of interest across time points was evaluated against human-expert segmented regions in both time points. Even though we found many radiomics features and preprocessing combinations with a high repeatability (Intraclass Correlation Coefficient (ICC) > 0.85), our results indicate that overall the repeatability is highly sensitive to the processing parameters (under certain configurations, it can be below 0.0). Image normalization, using a variety of approaches considered, did not result in consistent improvements in repeatability. There was also no consistent improvement of repeatability through the use of pre-filtering options, or by using image registration between timepoints to improve consistency of the region of interest localization. Based on these results we urge caution when interpreting radiomics features and advise paying close attention to the processing configuration details of reported results. Furthermore, we advocate reporting all processing details in radiomics studies and strongly recommend making the implementation available.

# Contents

# Introduction

The field of Radiomics is concerned with the extraction of quantitative imaging features to convert images into a large scale mineable data[1]. Lambin et al.[2] state the Radiomics hypothesis "that advanced image analysis on conventional and novel medical imaging could capture additional information not currently used, and [...] that genomic and proteomics patterns can be expressed in terms of macroscopic image-based features." The prognostic and discriminative power of radiomics features has been explored in cancer imaging with promising results[3–13] (including tumor locations prostate, lung, head and neck, brain, breast, glioblastoma, etc.).

Prostate cancer is one of the emerging applications with a strong need for improved characterization of the disease using imaging, as is evident from the ongoing efforts to standardize acquisition and reporting of the imaging findings[14,15]. Multiparametric MRI (mpMRI) is a well-established clinical tool used effectively for image characterization, treatment planning and response assessment. However, applications of quantitative analysis of mpMRI are very limited in the clinic.

The generally accepted standard of care is to use the Prostate Imaging Reporting and Data System (PI-RADS)[15], which establishes the guidelines for qualitative interpretation of mpMRI. In the research applications, most of the studies investigating quantitative analysis of mpMRI utilize basic imaging-derived features such as lesion volume[16], summary statistics of the Apparent Diffusion Coefficient (ADC)[17] and pharmacokinetic maps estimated from DCE MRI[18]. More recently, early results suggest that radiomics may have a role in differentiating non-cancerous prostate tissue from cancer, as well as grading prostate cancer[19]. Fehr et al.[4] combined a set of first and second order texture features computed in ADC and T2-weighted (T2w) in an automatic classification algorithm. They report that they could "obtain reasonably accurate classification of Gleason patterns." Wibmer et al.[5] also demonstrated that second order texture features on ADC and T2w may differentiate between cancer and normal tissue. They also found a correlation between ADC Entropy and Energy and Gleason score, but no correlation between T2w texture features and Gleason score. In another study Peng et al. [4] found that "[t]he combination of 10th percentile ADC, average ADC, and T2-weighted skewness with CAD is promising in the differentiation of prostate cancer from normal tissue. ADC image features and $K^{trans}$ moderately correlate with GS."

The premise of radiomics is that quantitative image features can serve as a biomarker characterizing the disease, or allowing prediction of response and thus providing decision support for patient management. To reliably derive conclusions based on any biomarker, a basic requirement is that its value must remain stable between two scans, if the conditions remain stable[20–23]. We use the the term "repeatability" for this attribute of a radiomics feature (others refer to it as "reproducibility"[20,21,24,25], or "stability"[3,22,23]). Good repeatability is a necessary, but not a sufficient condition for a high predictive power of a feature, meaning that if a feature has a high predictive power, its repeatability must be good. If a feature has a low repeatability, its predictive power must be low, too. But if a feature has a good repeatability, we cannot conclude anything about its predictive power. Gudmundsson et al.[23] demonstrated this aspect in their evaluations of feature importance and stability on different physiological time series.

Considering the repeatability of features is therefore a good measure for pre-selecting features for a classification task, given a large amount of features to select from. Such a selection is necessary since hundreds of feature sets are available for consideration in medical imaging[26]. This number multiplies if we consider different parameters, filters and preprocessing combinations.

Studies on the repeatability of features in medical images have been conducted with regard to various aspects. Zhao et al.[27] investigated repeatability of manual and computer aided diameter and volume measurements of lung lesions on CT test-retest. Others looked into the repeatability of MRI specific measures like per-voxel ADC[28] or quantitative parameters maps in T1 and T2*- weighted images[29]. Several studies have investigated a large set of radiomics features on CT lung cancer cases[3,20,21,24,25]. All of them found a large number of features with a good repeatability. Zhao et al.[20] found that many features are reproducible even under different CT reconstruction settings, but they also mention that repeatability of texture features is particularly susceptible to varying pre-processing schemes. Hu et al.[30] report 252 of 775 texture features have high repeatability on CT rectal cancer cases. They also state "that different filters have little effect to textural features." Another set of repeatability studies was conducted on non-small cell lung cancer cases[22,31,32]. While Leijenaar et al.[22] and van Velden et al.[31] report overall good repeatability, Desseroit et al.[32] mention a critical aspect: They found that "repeatability [...] varied greatly among metrics" and "repeatability also depended strongly on the quantization step, with different optimal choices for each modality". Critical issues were also raised e.g. by Emaminejad et al.[33] and Chalkidou et al.[34]. Emaminejad et al.[33] investigated various factors influencing texture feature calculation (Entropy) and found them "caus[ing] substantial variation". Chalkidou et al.[34] review 15 studies and "found insufficient evidence to support a relationship between PET or CT texture features and patient survival".

Based on these results, we feel that further emphasis on repeatability is needed in the radiomics literature. At the same time, radiomics analysis is fraught with complexities in identifying the optimal analysis parameters. As an example, we did not identify a consistent recommendation on how pre-filtering should be performed in PCa MRI radiomics characterization[4–6]. Depending on the specific study, image normalization (scaling and shifting) was applied only for texture features[5], for all features[4], or not used at all[6]. 3D computation of texture features is only mentioned in one study[5], while others[4,6] do not specify whether their computations were done in 2D or 3D. Overall, the description of the preprocessing often lacks details to allow for exact reproduction of the calculations.

Furthermore, most of the existing studies investigating radiomics feature repeatability focus on features extracted from CT. Radiomics analysis of MRI data poses significant challenges due to lack of signal normalization, more common acquisition artifacts, and lower spatial resolution. In their comprehensive review radiomics paper of 2016 Yip et al.[35] state that "the repeatability of MR-based radiomic features has not been investigated" and that "[u]nderstanding the stability of MR-based radiomic features between test and re-test scans can help identifying reliable features for radiomic applications, and thus would be a valuable future study." We still observe this gap in the present understanding of the feature repeatability applied to MRI analysis in general, and in prostate cancer imaging specifically.

In this study we assess the repeatability of radiomics features for small prostate tumors in multiparametric prostate MR images (mpMRI). We consider all features provided by the open source pyradiomics package[36], which are implemented according to consensus definitions of the Imaging Biomarkers Standardization Initiative (IBSI)[26]. We investigate various factors likely to influence the repeatability of features, such as: image normalization, 2D/3D texture computation, discretization with different bin widths, and image pre-filtering. Furthermore, repeatability also depends on the accuracy of tumor segmentation at both time points but we did not fully investigate the impact of that factor in this study. In our reporting we focus on disclosing all configuration details and make our implementation and data available. We note that we do not report results for all of the observations (all combinations of image types, regions, choices of normalization, pre-filtering and feature sets). Instead, our goal is to summarize findings of most relevance. This study is an extension of our previous work, where we evaluated the repeatability of volume and apparent diffusion coefficient (ADC)[37]. Those basic imaging features are widely recognized as valuable markers of prostate cancer[16,17,38].

## Methods

*Image Data and Segmentations*

The study used a previously published prostate mpMRI test-retest dataset composed of fifteen treatment-naïve men with biopsy-confirmed (n=11) or suspected (n=4) prostate cancer (PCa). Patients underwent a second MR within two weeks after the first MR, without any interim treatment[37]. The MRI exam included T2-weighted axial (T2w) images (TR 3350-5109 ms, TE 84-107 ms, FOV 140-200 mm), Subtraction images (SUB) calculated as the pixel-wise difference between the early post-contrast image phase and the pre-contrast phase of the Dynamic Contrast Enhanced (DCE) MRI acquisition, and Apparent Diffusion Coefficient (ADC) maps derived from Diffusion-weighted MRI (b=1400 s/mm^2, TR 2500-8150 ms, TE 76-80 ms, FOV 160-280 mm). See Fig. 1 - Fig. 5 for examples of the data used in this study.

A radiologist with 10+ years of experience in prostate mpMRI reviewed all of the images for the individual MR studies using 3D Slicer software (http://slicer.org)[39]. Regions of interest (ROIs) annotated included the suspected tumor, entire peripheral zone of the prostate gland, and the entire prostate gland identified in the baseline and follow-up T2w, SUB and ADC images. Image annotation utilized a visualization protocol whereas all of the individual images for a given patient and time point combination were shown to the reader for a single time point. However, the individual timepoints were randomized so that while annotating a given study the reader was blinded to the second time point for the same patient. Notably, all resulting tumor ROIs used for calculating the features were smaller than 0.8 ml. Details relating to the acquisition and annotation of the dataset are described in[37].

Upon review of the data, one of the subjects was excluded from the analysis of ADC features, since one time point had a significantly worse image quality (see Fig. 5). As a result, analysis of repeatability of the radiomics features for the ADC images was conducted using the data from the remaining 14 subjects.

Stability of the features is affected by the consistency of the segmentation of the region of interest between the baseline and repeat scans. In absence of the ground truth, we cannot evaluate the absolute accuracy of the manual segmentation performed by the radiologist. To mitigate the effect of potentially inconsistent segmentation of the region, we performed deformable intensity-based image registration[40] between the baseline and repeat T2w scans. We then visually evaluated the quality of the segmentation, and used the resulting transformation to transfer the segmentations from the first time point to the second. This automatically re-identified ROI was used to study whether repeatability is improved as compared to the calculations done using the manually defined regions.

*Feature Extraction*

Features were extracted for all ROIs using pyradiomics[1*], presented earlier in[36]. We extracted features from five feature classes[2*]: First Order, Shape, Gray Level Co-occurrence Matrix (GLCM), Gray Level Size Zone Matrix (GLSZM), and Gray Level Run Length Matrix (GLRLM) features. All features of each of these classes were extracted with the following exceptions: *Compactness1* and *Compactness2*, as well as

---

[1*] We used pyradiomics version v1.3.0.post59+g2e6d2c1 available through http://radiomics.io.
[2*] Throughout the text, where they correspond to implementations in pyradiomics, feature classes and preprocessing filters are capitalized and feature names are capitalized and denoted in *italics*.

*SphericalDisproportion* from Shape features were excluded because they are directly correlated to *Sphericity* (based on the definition of the feature, as discussed in the documentation of pyradiomics). *Flatness* and *LeastAxis* from Shape features were excluded because some tumor ROIs are only defined on one slice and these feature do not yield useful values for non-3D objects. *SumAverage* was excluded because it is directly correlated with *JointAverage*. *Homogeneity1* and *Homogeneity2* were disabled because they are directly correlated to *InverseDifferenceMoment*.

Pyradiomics allows preprocessing of (applying filtering to) the original image before feature extraction and offers the following options[36]: Original - leave the image unchanged, LoG - apply a Laplacian of Gaussian filter, Wavelet - apply Wavelet filters (all combinations of high- and low-pass filters on each image dimension), Square - take square of original image intensity, Square Root - take square root of absolute image intensity, Logarithm - take logarithm of absolute intensity + 1, and Exponential - takes the exponential of absolute image intensity. The last four filters also scale the values back to the original image range and restore negative sign if original was negative. For the LoG preprocessing we choose kernel sizes (sigmas) 1.0, 2.0, 3.0, 4.0, and 5.0 mm. Each feature was computed with each of the above mentioned preprocessing steps separately. We note that image pre-processing prior to feature calculation is currently not covered by the IBSI radiomics standardization initiative[26]. Here we consider all standard pre-processing approaches implemented in pyradiomics, which include pre-processing filters (i.e., wavelet and LoG) that have been shown to result in highly predictive feature sets[3,41].

Bias correction was applied to all T2w images to compensate for intensity non-uniformities using N4 Bias Correction[42] implemented in 3D Slicer[39].

MR image intensity is usually relative and not directly comparable between images. To reduce this effect we apply normalization. To test the effect of normalization on feature repeatability we included features computed with and without normalization. Basic normalization was performed by scaling and shifting the values of the whole image to a mean signal value of 300 and a standard deviation of 100. This means we would expect most of the values in the range of 0-600, assuming normal distribution of the intensities within the image. We also considered normalization based on a biologically comparable reference tissue region, which is assumed to be stable across time points and patients. For this we selected an ROI in a muscle region. The shifting and scaling factor for all image voxels was determined such that the mean signal in the reference ROI changed to 100 and the standard deviation to 10. The smaller range and lower mean for the latter normalization is due to the fact that the reference region only represents a small portion of the whole image intensity range and that muscle tissue has low intensity. This way the intensities of the whole image are keep in a reasonable range after this type of normalization. For the remainder of this paper, when we write about "normalization" we refer to the basic normalization, unless specifically mentioned otherwise.

Computation of texture features requires discretization (binning) of the image intensities into a limited number of grey levels. This can be done using either fixed number of bins, or the fixed bin size (see section 2.7 of the IBSI guidelines v6[26]). We used the latter discretization approach, as implemented in pyradiomics. Leijenaar et al.[43] argue why it is imperative to use a fixed bin width and not a fixed number of bins for discretisation. Additionally, according to Tixier et al.[44] the total number of bins for texture feature computation should be between 8 and 128. If we consider our value range of 0-600 after normalization, we should select bin widths between 5 and 75. However, for non-normalized images as well as normalization based on a reference ROI the intensity ranges vary and can be much larger. Hence, we selected bin widths 10, 15, 20, and 40 for performing our experiments, which will keep the number of bins below 128 for intensity ranges of up to 5120. We note that image binning as implemented in pyradiomics is based only on the intensities that are within the region of interest.

Texture features are computed as various statistics over specific matrices (e.g., co-occurrence matrix for GLCM). The dimensionality of the texture matrix defines the neighborhood (2D vs 3D) over which feature calculation is performed (see the pyradiomics documentation for more details). Since the choice between 2D- and 3D-based calculation of texture features is not obvious, and no comparisons of the two were done before, we included the comparison of stability for the two approaches in our study.

For all other configuration parameters of pyradiomics feature extraction we used the default settings (see http://pyradiomics.readthedocs.io for further information).

*Measure of Repeatability*

As a measure of repeatability we report the intraclass correlation coefficient ICC(1,1)[45]. The ICC considers the variation between repeated scans on the same subject in relation to the total variability in the population[46]. For our test-retest scenario with two time points it is defined as follows:

$$ICC(1,1) = \frac{BMS - WMS}{BMS + WMS},$$

where BMS is the between-subjects mean squares and WMS the within-subjects mean squares[45]. Hence BMS is an estimate for the variance between patients in our study and WMS an estimate for the variance over repeated measurements on the same patient.

The ICC is invariant with respect to linear scaling and shifting. This is a necessary property for using it to compare repeatability of features which operate in different unit and scale spaces. Since fixed thresholds for interpreting the ICC are problematic (see our Discussion section and e.g. Raunig et al.[46]), ICCs of different radiomics features should be compared to a reference within the study population. We use the *Volume* ICC as reference. Tumor volume is an important quantitative measure characterizing PCa, which has been investigated earlier[16], and evaluation of its repeatability in this specific dataset has already been presented in our earlier work[37].

*Evaluation Approach*

We start by evaluating the repeatability of a small subset of features, which were shown by others to perform well in PCa mpMRI radiomics-style analyses[4–6]. Specifically, this initial step of the evaluation focused on the following Intensity and GLCM radiomics features: *Volume*, *Entropy*, *Energy*, *Idm* (Inverse Difference Moment), *Correlation*, *Contrast*, *Variance*, *Skewness*, *Median*, *Mean*, *Kurtosis*, and *10Percentile* (10th percentile of intensity distribution). Using this relatively small dataset, we first explore the effects of normalization, intensity bin size, and pre-filtering on the repeatability of those features for all image types (ADC, SUB, T2w). Based on the observations on this reduced feature set, we aim to identify preprocessing and feature extraction parameters that lead to improved repeatability, and continue with the evaluation of the complete radiomics feature set using the selected processing options. Since it is unfeasible to look into all features individually, in this phase we focus either on summarizing statistics over all features or on a selection of the top 3 best performing features per feature group.

The large amount of data generated by our extraction also does not allow us to explore all aspects in detail in this paper. The supplementary material and source code accompanying the manuscript include investigation of additional aspects of the dataset that were not included in the manuscript.

*Data Availability*

The MRI datasets on which is study is based are from the QIN-PROSTATE-Repeatability TCIA collection.

The extracted features used for all analyses in this paper are located at:
https://github.com/michaelschwier/QIN-PROSTATE-Repeatability-Radiomics/tree/master/EvalData

*Code Availability*

Pyradiomics is available at:
https://github.com/Radiomics/pyradiomics (*v1.3.0.post59+g2e6d2c1*)

Code to perform the analysis and figure generation is available at:
https://github.com/michaelschwier/QIN-PROSTATE-Repeatability-Radiomics
This repository also contains example code on how the features were extracted from the image data.

# Results

To investigate the repeatability of radiomics features for small prostate tumors in mpMRI we analyzed a large set of features under various preprocessing combinations. As there are numerous combinations of image types, regions, choices of normalization, pre-filtering and feature sets, we selected the most relevant findings in our results.

*Selected Features*

**Normalized vs Non-Normalized Images**

Our evaluation of a small subset of features that were shown by others to perform well in PCa mpMRI analysis yielded the following results: For the Tumor ROI in normalized ADC images (see Fig. 6a) *Entropy*, *Idm* (Inverse Difference Moment), *Correlation*, *Median*, *Mean*, and *10Percentile* (10th percentile of the intensity distribution) reach ICCs equal or better than *Volume* (ADC *Volume* ICC=0.7). In the Peripheral Zone in normalized ADC images (see Fig. 6b) *Entropy*, *Energy*, *Idm*, *Contrast*, *Median*, *Mean*, and *10Percentile* reach ICCs around 0.91 or higher, performing better than Volume (ADC *Volume* ICC=0.85). Looking at the ADC ICCs in the Whole Gland (see Fig. 6c) we observe that no feature reaches the 0.99 ICC of *Volume*. Normalization leads to improved ICC in ADC images in most cases (see Fig. 6a-c). Notable exceptions are *Variance* in the Peripheral Zone as well as *Entropy*, *Energy*, and *Correlation* in the Tumor ROI. For these exceptions the difference to the ICC of the corresponding normalized feature was always smaller than 0.1. Note that *Skewness* (measure of asymmetry of the distribution about the mean) and *Kurtosis* (measure of peakedness of the distribution) are by definition not influenced by normalization. However, they also never reach the reference *Volume* ICC.

In the Tumor ROI in SUB images (see Fig. 7a) the Volume ICC is 0.57. *Entropy*, *Idm*, *Contrast*, *Median*, *Mean*, and *10Percentile* perform better with *Entropy*, *Idm*, and *10Percentile* being the best, reaching ICCs of 0.83, 0.91, and 0.87 respectively. In the Peripheral Zone in SUB images the reference *Volume* ICC is 0.51 and is surpassed by the following features (see Fig. 7b): *Entropy*, *Idm*, *Correlation*, *Contrast*, *Variance*, *Median*, *Mean*, and *10Percentile*. Except for *Variance*, all of these reach ICCs of 0.63 or higher (up to 0.85 for *Idm*). For SUB Whole Gland no feature reaches the 0.94 ICC of *Volume* (see Fig. 7c). Regarding normalization there is no clear trend on SUB images. Depending on the specific structure and feature, normalization may or may not lead to improved ICC. For example, the ICC of *10Percentile* is higher on normalized SUB images in the Whole Gland and the Peripheral Zone, but lower ICC in the Tumor ROI. The ICC of *Mean* is higher on normalized images in the Peripheral Zone but in the other structures normalization has the opposite effect.

On T2w images neither in the Peripheral Zone, nor in the Whole Gland does an ICC of any feature reach the *Volume* ICC (see Fig. 8b-c), which is 0.86 for the Peripheral Zone and 0.95 for the Whole Gland. Only in the Tumor ROI *Entropy*, *Energy*, and *Variance* reach an ICC higher than the reference T2w Volume ICC of 0.86 (see Fig. 8c). Contrary to ADC images, normalization leads to lower ICCs in T2w images. The only exception is *Correlation*, which has a higher ICC for the Whole Gland when normalized.

In the following we report further results based on normalized ADC and non-normalized T2w images only, since results in this section showed better overall repeatability under these configurations.

**Influence of Different Bin-widths**

Fig. 6, Fig. 7, and Fig. 8 indicate that different bin widths do not result in very strong variations of the ICC for most features. The kernel density estimation (KDE)[47,48] plot in Fig. 9a supports this observation. It shows the distribution of the maximum difference between highest and lowest ICC per feature depending on bin width for the selected features. In Fig. 9b we additionally plotted the same distribution for all GLCM features and all pre-filtering options. For the majority of features the maximum difference is around 0.2.

Another insight into the influence of bin width is given in Fig. 10. We ranked the ICC for each feature depending on the bin width and plotted the rank distribution. We can see that for the lowest and highest bin widths, the best and worst ranks are appearing about equally often. Bin width 15 and 20 cover the middle ranks.

In the following we report all results based on bin width 15, since the results in this section indicate that this yields a reasonable average estimate of the feature repeatability.

**Influence of Pre-filtering**

Fig. 11 illustrates the influence of 2D and 3D wavelet pre-filtering and texture extraction in T2w Tumor ROI images for the selected features. In most cases different wavelet filters cause strong differences in ICCs for the same features and those ICCs are spread between very low and very high values. Only *JointEntropy* has consistently high repeatability for all wavelet pre-filters as well as for 2D and 3D texture extraction.

For some features like *Contrast*, wich did not show good performance without pre-filtering, certain filters improve the repeatability to reach an ICC above the reference threshold. However, among all ICCs above the *Volume* ICC reference threshold, no clear trend towards a certain wavelet filter type can be observed. Filter types that improve repeatability for one feature may lead to poor results for other features. Consider for example the effect of applying preprocessing using the Wavelet-LL filter on *JointEnergy* and *JointEntropy* versus the same filter on *Correlation* and *Contrast*.

Similar results can be observed for Laplacian of Gaussian (LoG) and Single Pixel pre-filtering (filters that do not consider the pixel neighborhood) in T2w Tumor ROI images (see Fig. 12 and Fig. 13 respectively). Again the ICCs are spread over the whole range from good to bad. Pre-filtering did not lead to high ICCs with either 2D or 3D feature extraction. However, a few filters tend to consistently result in low ICC values, for example LoG with a high sigma, and exponential pre-filtering in combination with 3D feature extraction.

*Top 3 Features per Feature Group*

Given all possible combinations of pre-filtering options, we selected the 3 most repeatable features for each of the feature classes implemented in pyradiomics (namely Shape, First Order, GLCM, GLSZM and GLRLM). We then investigated whether any specific pre-filtering approach consistently resulted in improved repeatability of these selected features. For these top 3 features Fig. 14 illustrates the range of ICCs under all pre-filtering options for the Tumor ROI and Peripheral Zone in T2w images. We observe that other shape features have a better repeatability than Volume. They are also by definition not influenced by any pre-filtering. The ICCs of other features are again spread over a wide range. In the Tumor ROI (Fig. 14a) no single filter appears to result in consistently more stable features. However, several filters consistently result in an ICC below Volume ICC (e.g., Logarithm and Exponential). In the Peripheral Zone, only a few wavelet filters yield ICCs above the reference for all top 3 features with the exception of GLCM *ClusterProminence*, for which also Logarithm filtering reaches a higher ICC. On the low end particularly the Exponential filter performs consistently weak.

Fig. 15 illustrates the same top 3 analysis on ADC images for Tumor ROI and Peripheral Zone. In the Tumor ROI the ICCs are spread over a wide range again (see Fig. 15a). Even though many filters are related to high ICCs on several features, no consistent trend can be observed. A few features have a strong tendency towards yielding high ICCs, like LoG sigma 3.0 mm, but we can always find an exception. In the Peripheral Zone most filters are associated with ICCs above the reference for almost all features (see Fig. 15b). Also the spread of ICCs depending on the pre-filter is much smaller with a few exceptions (e.g. GLSZM *SmallAreaEmphasis*, or Firstorder *90Percentile*).

*Overview of Pre-filter Performance over all Features*

For an overview of the influence of pre-filtering options on repeatability of all features we considered all features and pre-filter combinations on T2w and ADC images in the Tumor ROI and Peripheral Zone. We selected all of those combinations that had ICC above the corresponding Volume ICC. The plots in Fig. 16 and Fig. 17 show how often a particular filter appears among those for Tumor ROI and Peripheral Zone in T2w and ADC images. On T2w Tumor ROI Wavelet-HH, Wavelet LH, and smaller sigma LoG filters are most prominent. Also the other Wavelet filters as well as the original image (no filter applied) often lead to high ICC values. On T2w Peripheral Zone only Wavelet-HH, and Wavelet-LH stand out. In the Tumor ROI on ADC images Original and LoG filters perform well, while among the wavelet filters Wavelet-LH and Wavelet-LL stand out. Among the Single Pixel filters Square shows a strong performance. For ADC Peripheral Zone the filters corresponding to high ICCs are more equally distributed. The only exceptions are Exponential and Logarithm, which both are less often associated with good repeatability. Overall, no filter is consistently related to high ICC values - neither on T2w nor on ADC images (see Fig. 16 and Fig. 17 for details).

*Normalization against a Reference Region*

Normalization against a reference region on T2w results in a strong decline of repeatability for most of the selected features (see Fig. 18a). We also again selected the top 3 most repeatable features per feature class for the original T2w as well as the T2w that was normalized against a reference region. Fig. 18b shows that some features remain stable but repeatability of most original T2w top 3 features strongly declines on normalized T2w and vice versa.

*Re-identification of ROIs by Registration*

Fig. 19 illustrates the difference in ICCs between using the original ROIs versus ROIs which were transferred from on time point to another via image registration. Overall half of the features show an improvement in their ICC values while for the other half the repeatability declines.

## Discussion

Despite the small sample size and staged approach to conducting the evaluation, our study of radiomics feature repeatability resulted in a large number of observations. To a degree, this was caused by our attempts to find patterns and to be able to explain the results and make specific recommendations. In the following we discuss our interpretation of the main findings of the study.

Among the features reported in literature for prostate cancer analysis[4–6] in T2w and ADC MR images, we observed good repeatability (ICC ≥ Volume ICC) for *JointEntropy*, *Idm*, *Median*, *Mean*. However, for others we could not confirm good repeatability of features calculated over Tumor ROIs. Notable examples include *Kurtosis* and *Contrast* that in all cases are underperforming compared to the ICC of *Volume,* in some cases reaching values close to 0. Furthermore, even features with good repeatability showed these only under specific preprocessing configurations. On our data normalization led to improved ICCs on ADC images for most features, while for T2w images normalization did not result in ICC improvements. The observation that normalization improved repeatability for ADC is not completely unexpected. Although ADC is a quantitative measure that is supposed to be consistent across exames and platforms, variability of around 5% was observed even under perfect conditions for a temperature controlled phantom[49]. For SUB images (notably, not commonly utilized in the studies developing radiomics signatures for PCa[4–6]) *JointEntropy*, *Idm*, *Contrast*, *Median*, *Mean*, and *10Percentile* showed good repeatability, but requiring normalization for some and not for others. We can already see that it seems difficult to unequivocally recommend a combination of normalization choice and features that would yield a consistently good repeatability. This is even more clear if we look at the Peripheral Zone or Whole Gland.

As we saw in Fig. 9 and Fig. 10, even when the bin width is selected within the recommended limits[43,44], it still has an influence on the repeatability, although it is not too strong in most cases. Nevertheless, the differences we observed lead us to advice to evaluate the influence of bin width in any new study.

Our investigation of pre-filtering options and parameterization of the texture feature computation revealed further challenges in extracting repeatable radiomics signatures. We found that the use of pre-filtering for the literature reported features for Tumor ROI on T2w images introduces even more variation in the ICCs per feature across the various pre-filtering options (Fig. 11, Fig. 12, Fig. 13). This strong variation of ICCs is also confirmed if we look at the results for the top 3 best performing features in each feature class (Fig. 14 and Fig. 15). Notable exception is the relative stability of features in the Peripheral Zone ROI for the ADC images (Fig. 15b). Furthermore, our analysis of how often each filter is related to an ICC above *Volume* reference (Fig. 16 and Fig. 17) again reveals that we cannot suggest a filter that consistently improves repeatability.

Nevertheless, we can observe certain trends for filters that predominantly relate to below-reference ICCs. On T2w images these are all single pixel filters as well as large sigma LoG. On ADC images the results are less consistent. Logarithm and Exponential have a low repeatability in the Peripheral Zone, while for the Tumor ROI Logarithm, Wavelet-HH, and Wavelet-HL have the weakest performance. Hence, based on our results we would recommend to leave these filters out for the designated image types and structures.

However, in any case there are still many other filters which yield high ICCs, but not consistently enough to be able to pick a few for general recommendation. Even if we narrow down to a specific image and structure, we cannot single out a small set of pre-processing configurations which consistently result in improved ICCs. The results are too diverse. Depending on the feature, different preprocessing options yield the best repeatability.

There is an exception, however. Our results show several shape features (e.g., *SurfaceVolumeRatio* and different uni-dimensional diameter measurements) with a better repeatability than *Volume*. By definition these are also invariant under any of the investigated pre-processing options. Furthermore, these shape features are correlated to *Volume* and capture less information about shape than *Volume*. Hence most of these features are not likely to add any information that is not captured by *Volume* already.

Even though we could not find configurations which consistently improved the repeatability for all (or a large set of features), we still found many features, which - under certain configurations - have a better repeatability than our reference. For the Tumor ROI about 70 features are performing above par on T2w as well as on ADC (see Fig. 16 and Fig. 17). This could indicate that different features simply require different configurations. However, we don't believe this is the case, considering the small study size plus the fact that no obvious pattern emerges between certain pre-processing configurations and sub-groups of features. Furthermore, we were not able to fully reproduce the good repeatability of features reported in literature for PCa mpMRI on our dataset. This is another indication that there are many factors - even beyond the ones we assessed in this study - that influence the repeatability of features.

There are multiple factors that affect feature repeatability. An important one is the consistency in identification of the region of interest between the two timepoints. Even though the ROIs used in this study were contoured by an expert radiologist with 10+ years of experience in prostate MRI, we can expect that the segmentations will not be perfectly consistent between time points. Fig. 2 and Fig. 3 illustrate that it is not always clear if the segmented regions really completely represent the same tissue region. A perfect match of manually defined ROIs cannot be expected in practice. However, our attempts to automatically re-identify the ROI in the repeat scan for the T2w image produced mixed results. We did not observe a consistent improvement of repeatability (see Fig. 19). On the contrary, repeatability is decreased dramatically for some of the features. Most of the Tumor ROIs are small, and there is inherent error in any image registration registration result, which cannot be quantified automatically at an arbitrary location (see Fig. 4 for an example). This could explain our lack of clear improvement by using registration, which may imply registration errors are overall comparable with the inconsistencies of the manual annotations.

In our study, we made the decision to use ICC as the measure for evaluating repeatability. We use the ICC because it is commonly used in radiomics studies, and as such is a *de facto* convention in the field. It also has the advantages of being invariant with respect to linear scaling and shifting. Some authors indicate fixed thresholds for interpreting the ICC[22,30]. However, these can only be valid as a reference if our population variance was expected to be comparable to the ones used in other studies. In general, this is not the case though. As Raunig et al. note[46]: "ICC values for a very heterogeneous subject sample may yield very [sic] nearly perfect correlation based solely on the between-subject variance". Hence, ICCs of different radiomics features over the same population of subjects should be compared to a reference within this population. Considering those limitations of the absolute ICC threshold, we use the *Volume* ICC as the reference. Tumor volume is an important quantitative measure characterizing PCa, which has been investigated earlier[16], and evaluation of its repeatability in this specific dataset has already been presented in our earlier work[37].

Another measure which is commonly used to assess repeatability is the repeatability coefficient (RC)[46,50]. However, the RC is not invariant with respect to scaling and it is denoted in absolute units of the feature. If two features are not expressing their values in the same units and scale space, it is not valid to compare them based on their RC. Since most radiomics features are abstract measures with no direct real-world interpretation we cannot assume that they operate in the same unit and scales space and thus cannot not use the RC for comparing different features. The RC is rather designated for assessing the agreement of two methods, measuring the same quantity[50], and exposing the expected absolute differences between repeated measurements (limits of agreement).

Our study has limitations. A number of them are inherent to the dataset we used to extract the features (the limitations of the study that generated the MRI data and annotations are discussed extensively by Fedorov et al. in[37]): small sample size, small volumes of the identified tumors, lack of the analysis of multi-reader consistency. Specific to our analysis, we did not consider rigorous statistical modeling and testing for our evaluation in this study for several reasons. First, the sample size is rather small. Second, the intention of this paper is an overview of the effects of pre-processing variations on the repeatability of radiomics features. A thorough statistical analysis of all the variations considered in this paper would extend the scope. Since all data of this study is available, we hope this will encourage researchers to perform rigorous statistical analysis as an extension to this study.

## Conclusion

Our study shows that radiomics features, as evaluated on the specific PCa mpMRI dataset we used, vary greatly in their repeatability. Furthermore, repeatability of radiomics features evaluated using ICC is highly susceptible to the processing configuration. Even on our small study population, the results already show that the type of image, preprocessing, and region of interest used to evaluate the feature can vastly change the repeatability of certain features. This could contribute to the explanation of why feature recommendations among recent studies are not consistent, and why we could not confirm good repeatability for some of the literature reported features.

We suggest caution when utilizing prior studies as a basis for pre-selection of radiomics features to be used in radiomics signatures. Whenever possible, repeatability analysis on a representative dataset should be done as part of the study-specific feature selection procedure. If repeatability analysis is not possible or is impractical, and prior evidence is used for feature pre-selection, we recommend paying close attention to the reported configuration of the feature extraction process. Furthermore, it is important to consider whether the relevance of the assumptions on the reported study population are also valid for the planned study. Specifically, based on our observations, we find that prior studies may not provide sufficient detail about their feature calculation process to be meaningfully used to guide feature selection for new studies.

When publishing findings on radiomics studies (be it on repeatability or any other performance measure) we especially advocate reporting all details regarding the preprocessing. To increase reproducibility of study findings we also strongly recommend following the consensus definitions of features[26] and making the implementation available. When sharing of the dataset is not possible, we recommend that the pertinent details about the study population are reported to help in the reuse of conclusions of the study. As one example, it is common that distribution of the tumor volumes is not summarized, although it is widely recognized that size of the region of interest has strong effect on the measurements extracted. To support reusability and further investigation, our study utilized the publicly available open source radiomics library pyradiomics[36]. Furthermore the dataset used in this study and the calculated radiomics features are publicly available.

In conclusion, we were not able to determine a general set of universally stable feature and preprocessing recommendations. Nevertheless, we found many features with a considerably higher repeatability than our reference (*Volume*). Most prominently these include the top 3 features for each feature group. Based on the analysis of our data, these can be strong candidates for good prognostic features.

## Acknowledgement

Funding support: NIH U01 CA151261, U24 CA180918, P41 EB015898, R01 CA111288, R01 CA160902, U24 CA194354, and U01 CA190234.

# Figures

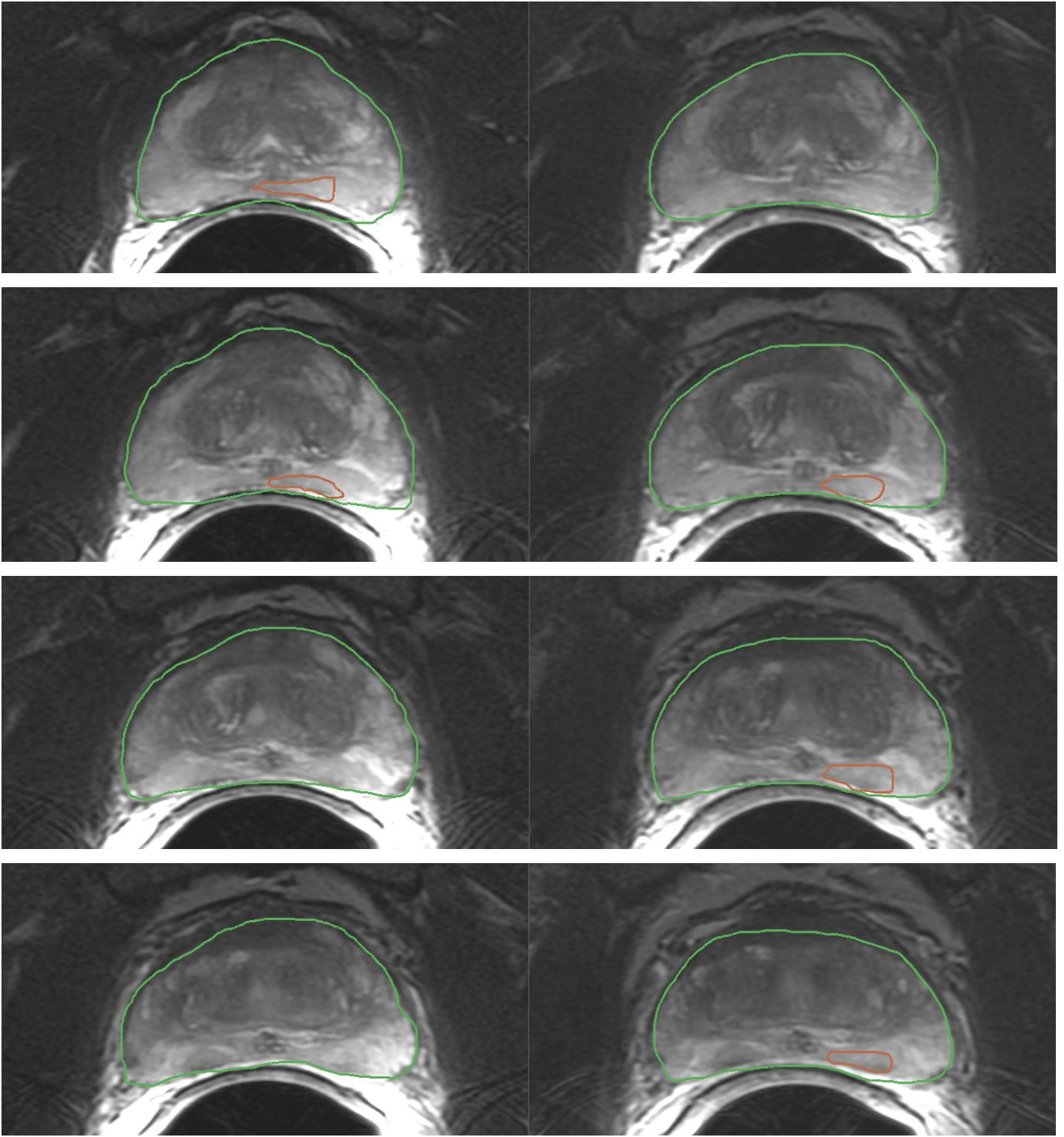

**Figure 1:** Segmentation of the Tumor ROI (red) and Whole Gland (green) on T2w images for case 2 time point 1 (left) and time point 2 (right). The picture shows 4 slices matched visually to show the same spatial location. Even though the disease is stable and the acquisition parameters are comparable, the location and appearance of the actual tumor region (and whole gland as well) changes between the timepoints.

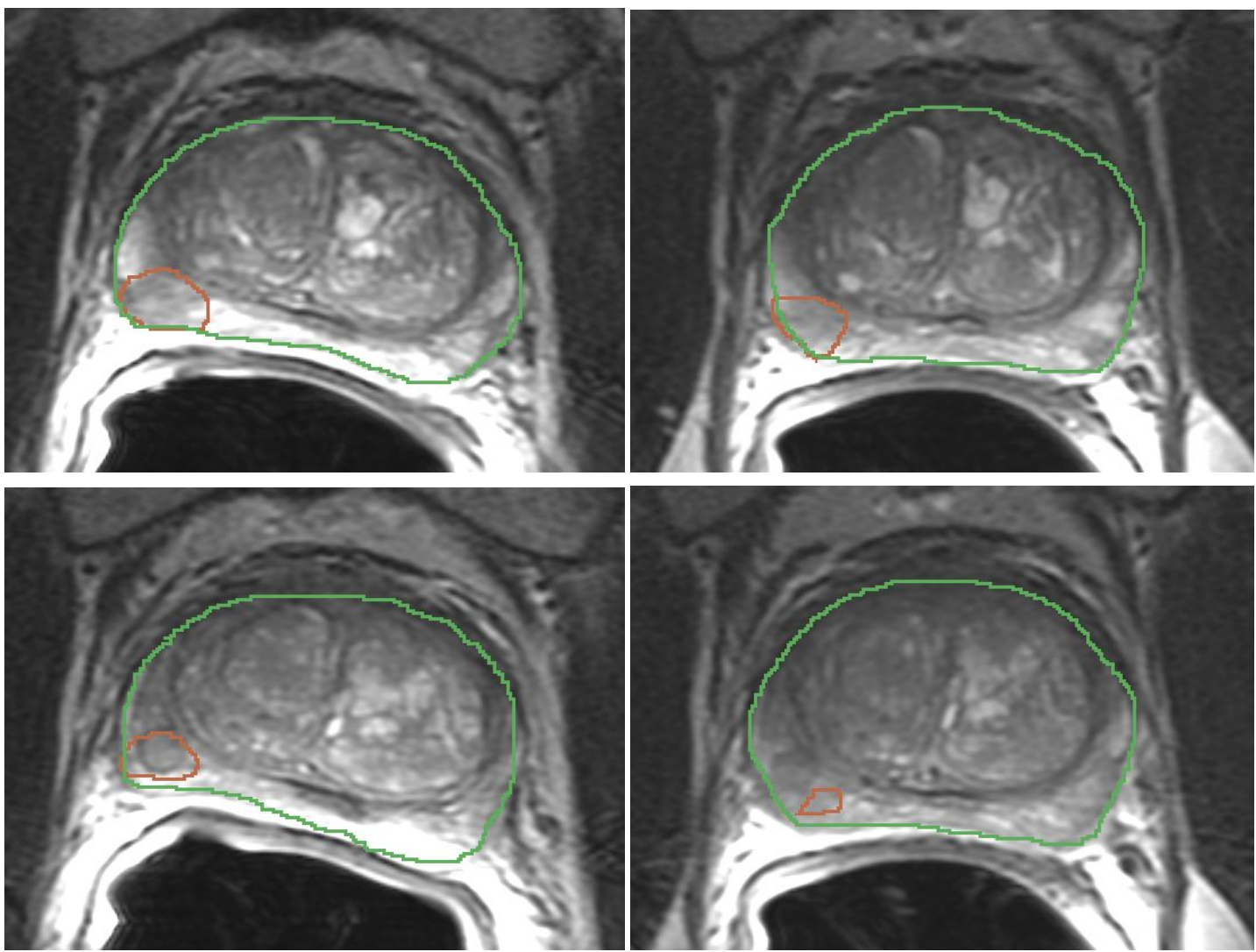

**Figure 2:** Segmentation of the Tumor ROI (red) and Whole Gland (green) in T2w images for case 7 time point 1 (left) and time point 2 (right). Each slice pair was selected to match location. The individual time points were segmented manually by the domain expert blinded to the second time point. Upon the review of the images corresponding to the time points side by side, it becomes apparent that the location of the segmented region is not consistent for the slice shown in the bottom row.

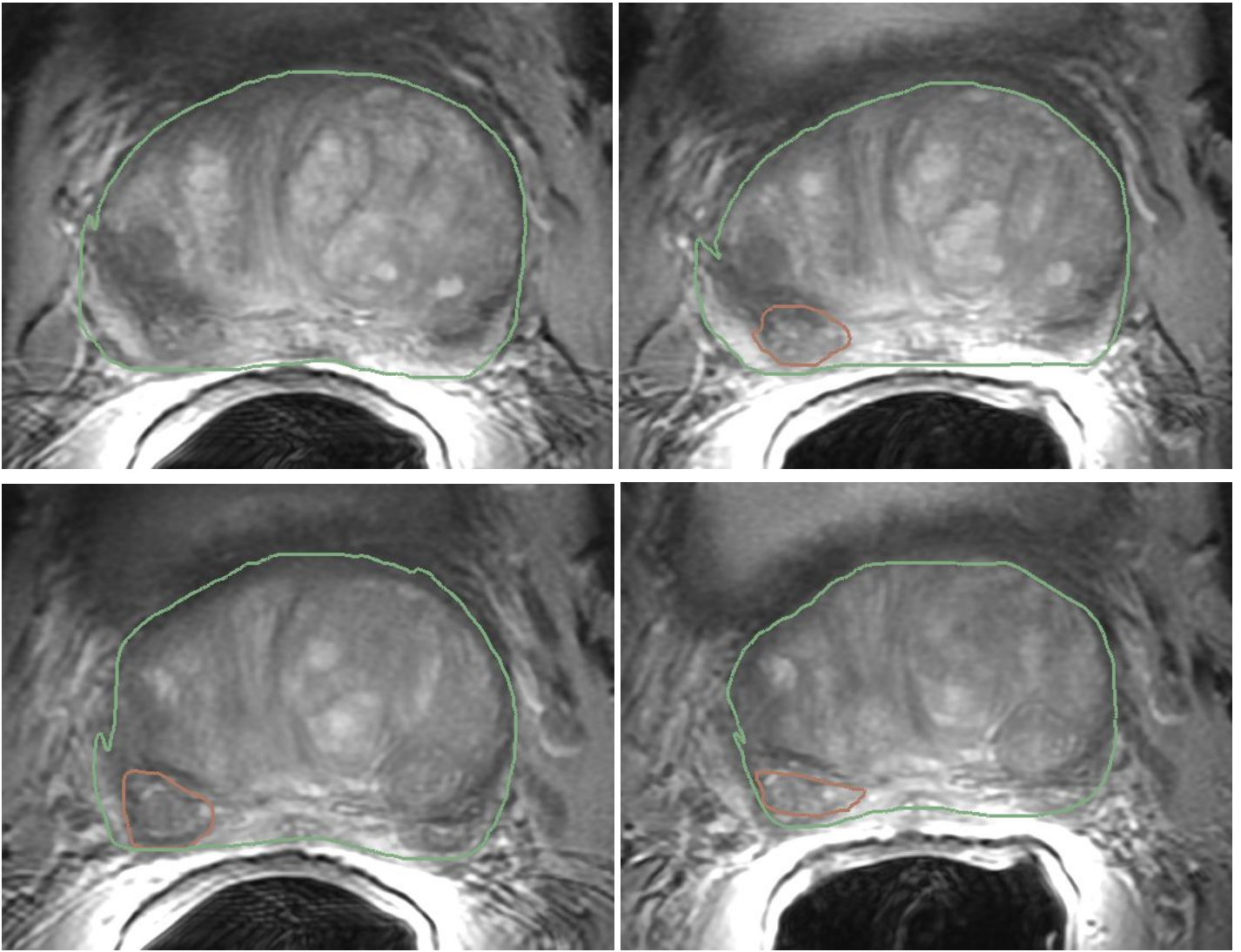

**Figure 3:** Segmentation of the Tumor ROI (red) and Whole Gland (green) on T2w images for case 9 time point 1 (left) and time point 2 (right). Each slice pair was selected to match location. The pictures show that there is a perceivable difference in the segmentations: the region corresponding to the localized lesion in time point 2 (top row) is not annotated in time point 1. Bottom row shows differences in the texture appearance within the lesion, in addition to potentially inconsistent boundary localization by the reader.

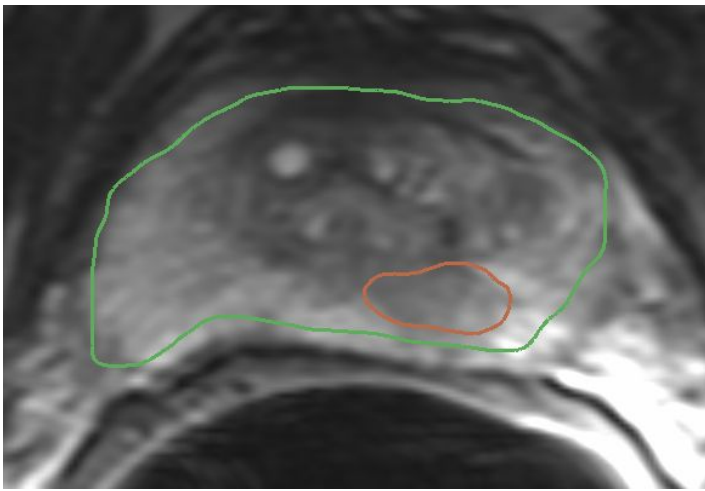 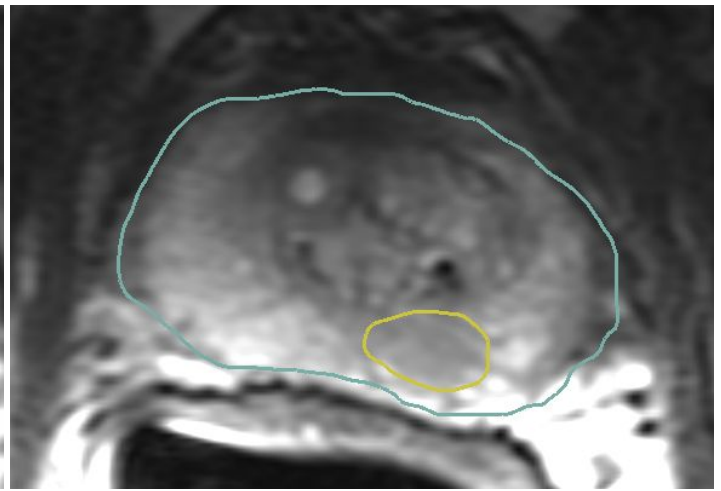

(a) Original segmentations time point 1

(b) Original segmentations time point 2

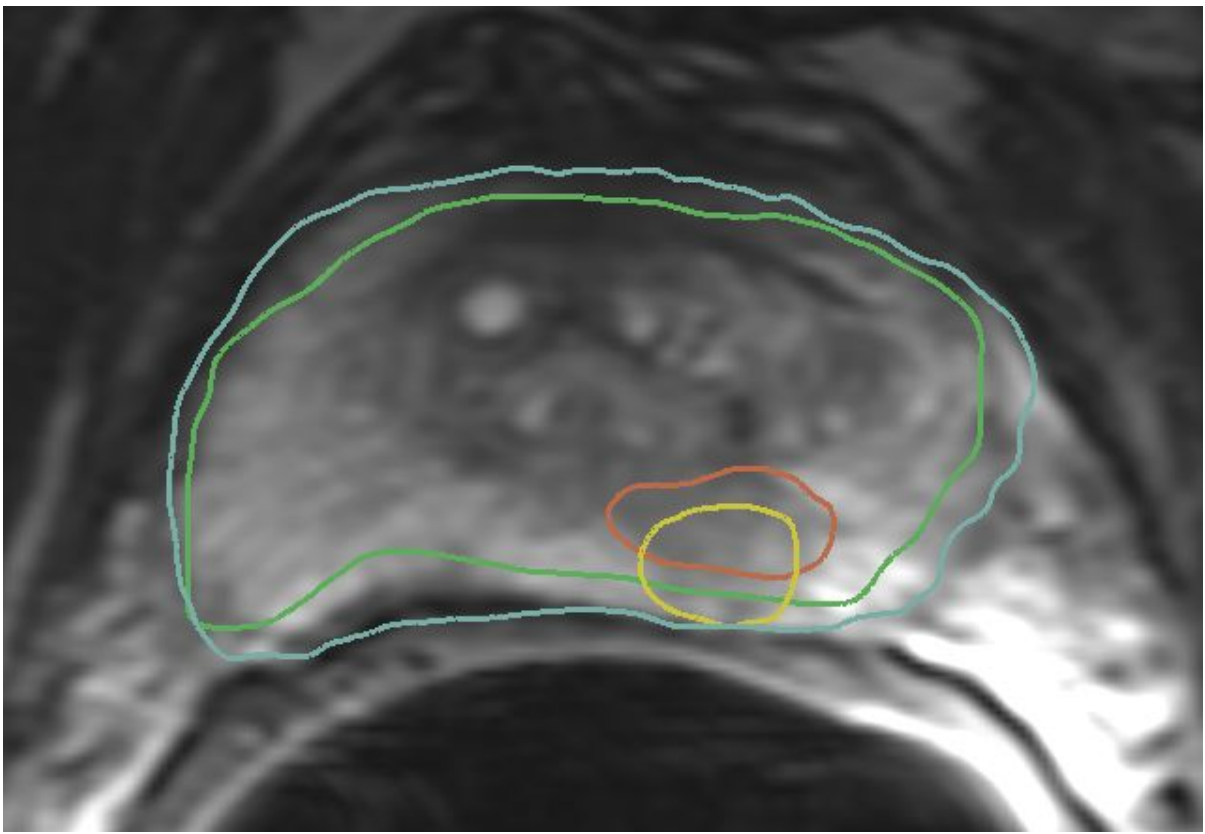

(c) Segmentations of time points 1 (original) and 2 (registered), overlayed on time point 1 image

**Figure 4:** Results of registering time point 2 onto time point 1 for case 3. Green: Whole Gland time point 1; blue: Whole Gland time point 2; red: Tumor ROI time point 1; yellow: Tumor ROI time point 2. The picture shows that after registration the segmentations are not aligned, which can be contributed to both the inconsistencies in segmentation of the corresponding structures and registration errors.

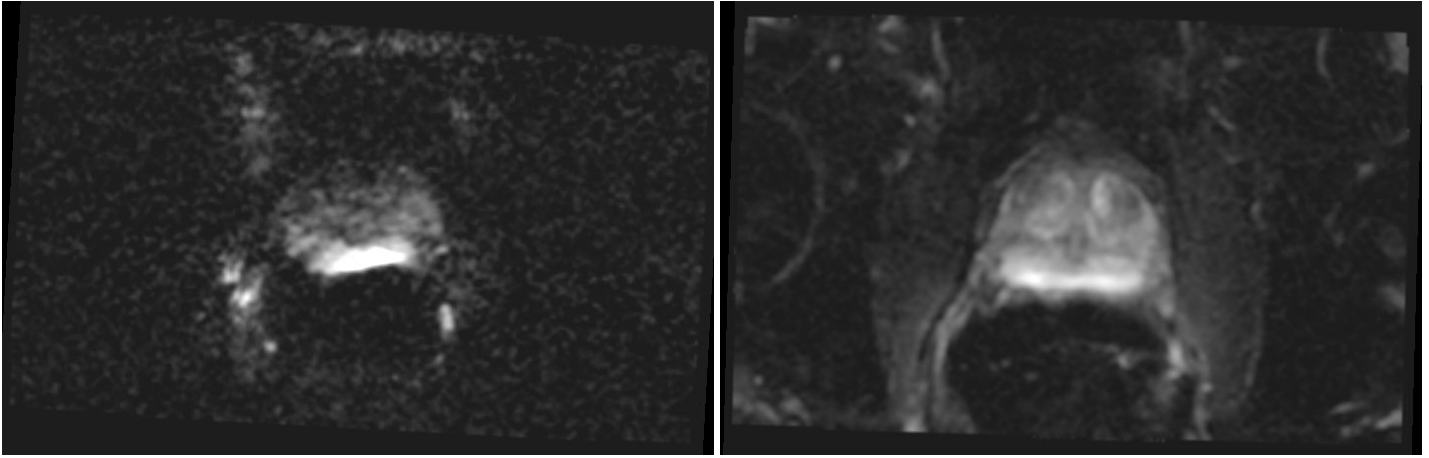

**Figure 5:** ADC case 1 time point 1 (left) and time point 2 (right). Visual appearance of images is very different, indicating a deviation in acquisition protocol. Hence, this case was removed from the study.

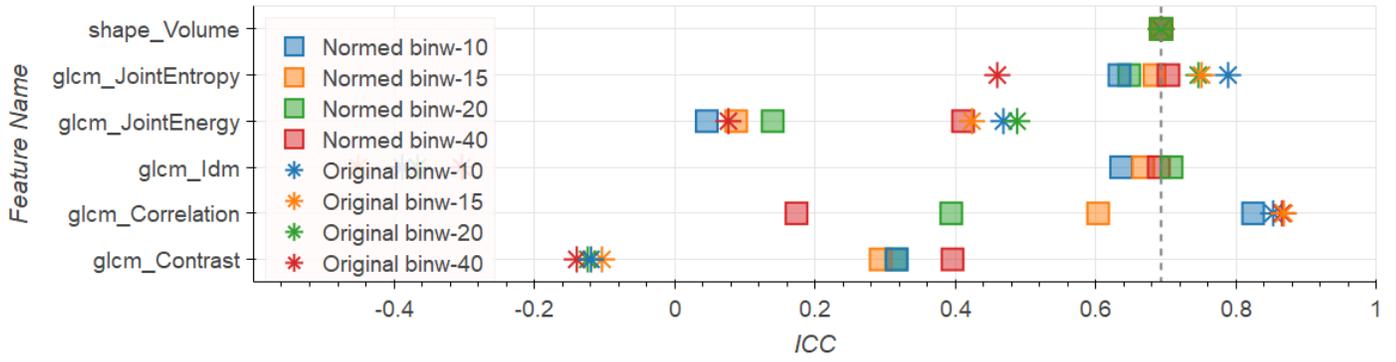
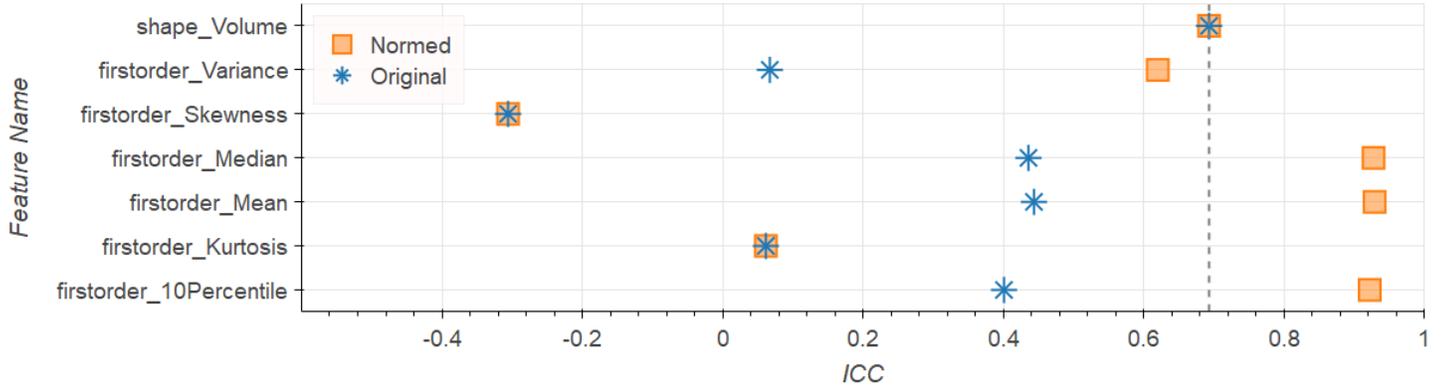

(a) Tumor ROI (top: texture features, bottom: first order features)

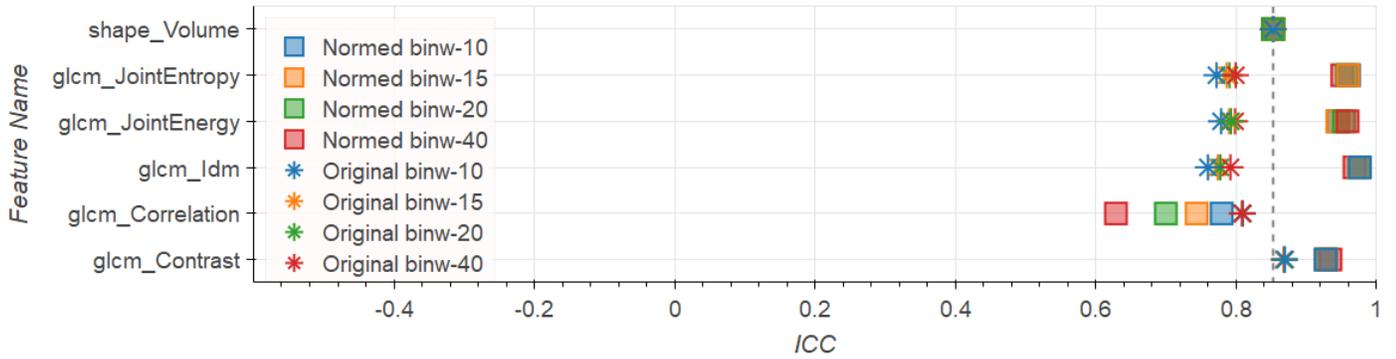
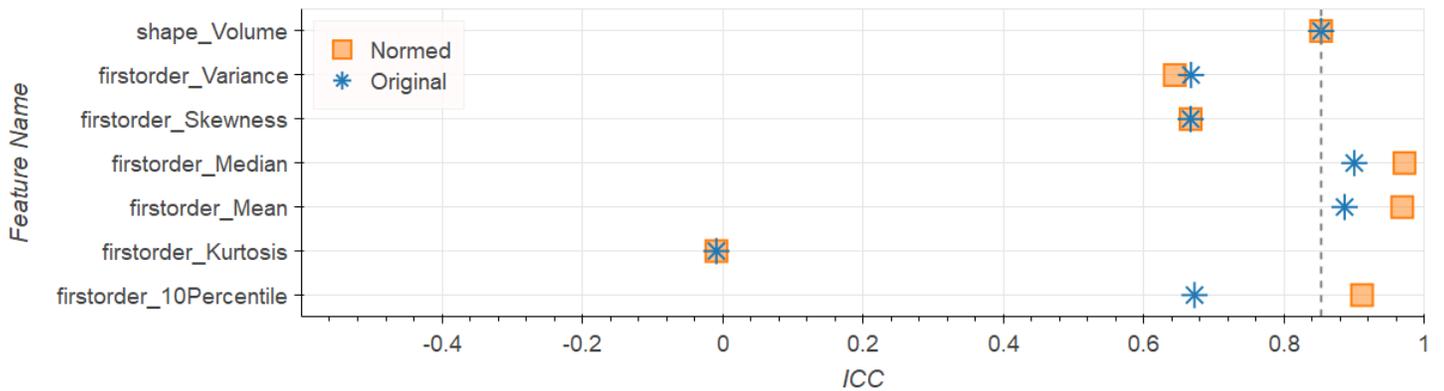

(b) Peripheral Zone (top: texture features, bottom: first order features)

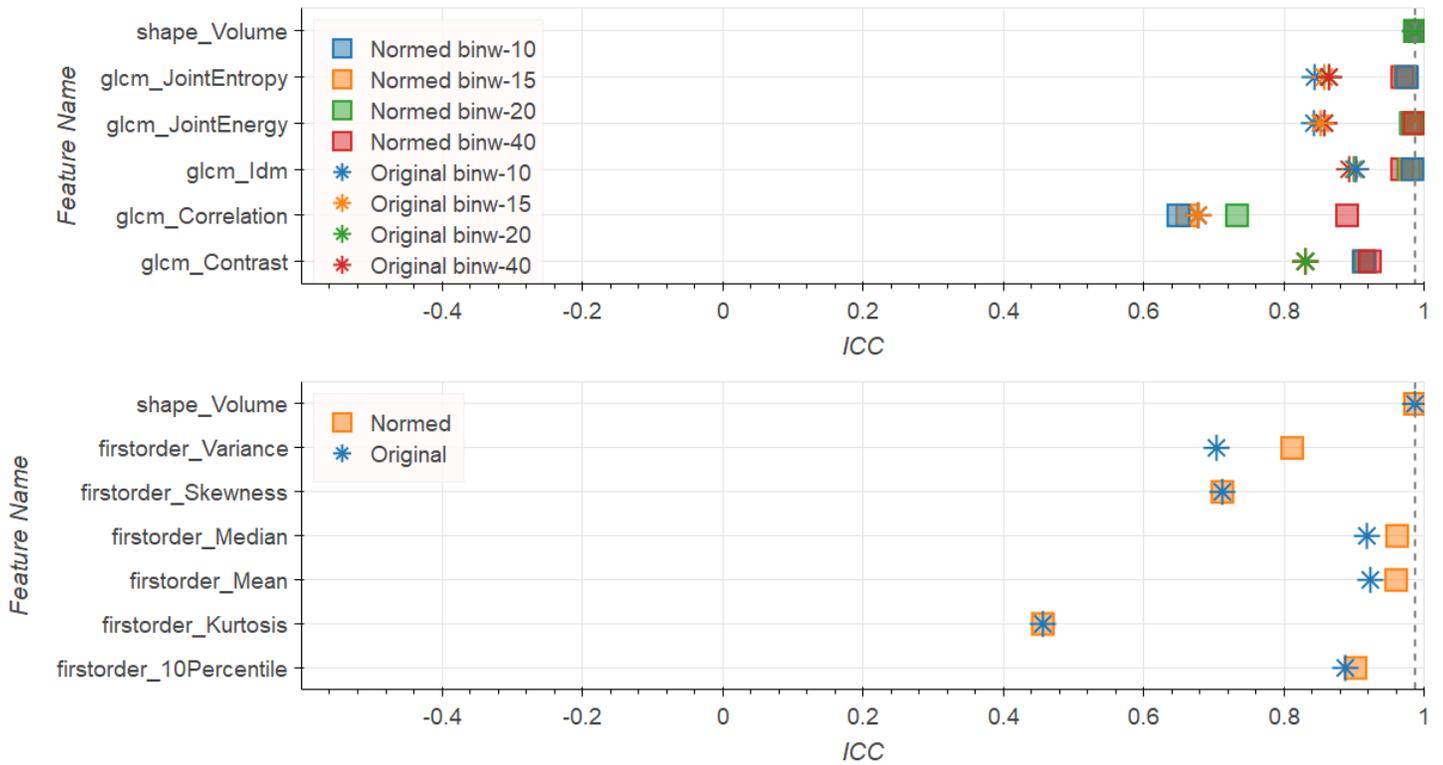

(c) Whole Gland (top: texture features, bottom: first order features)

**Figure 6:** ICC for selected features computed on different ROIs on ADC images. Texture features are computed in 2D. Colors represent the bin width for the texture computations, glyph shape represents if the image was normalized or not. No filtering was applied to the image. Dashed line indicates the reference *Volume* ICC. Results show that normalization rends to improve ICCs while bin width has only marginal influence.

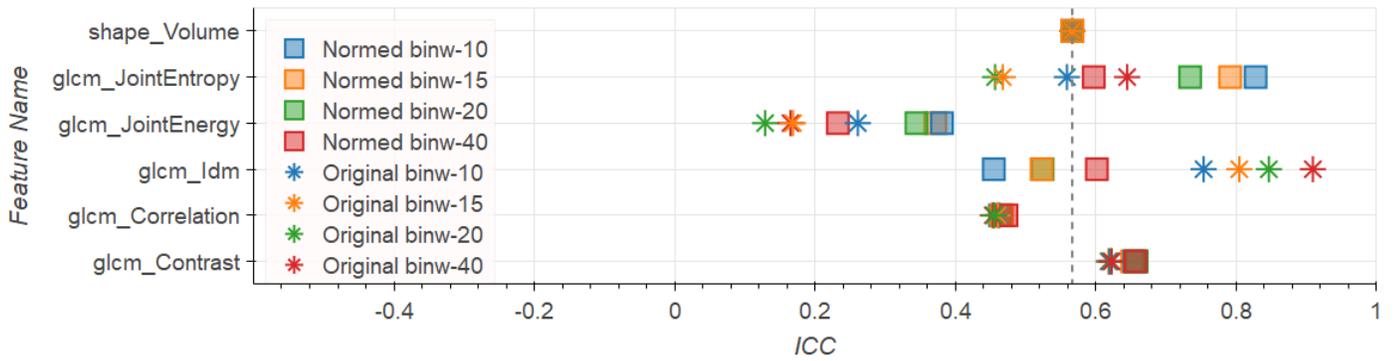
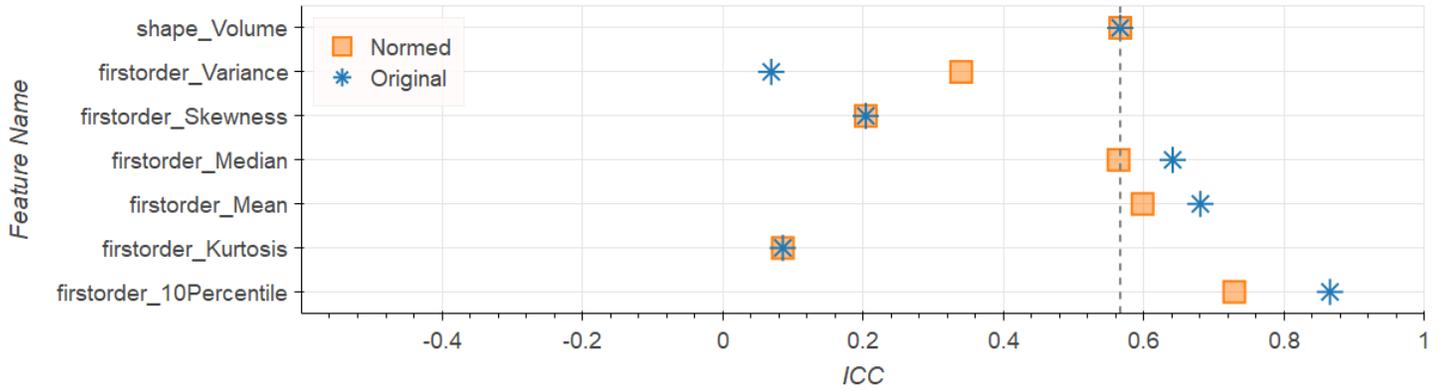

(a) Tumor ROI (top: texture features, bottom: first order features)

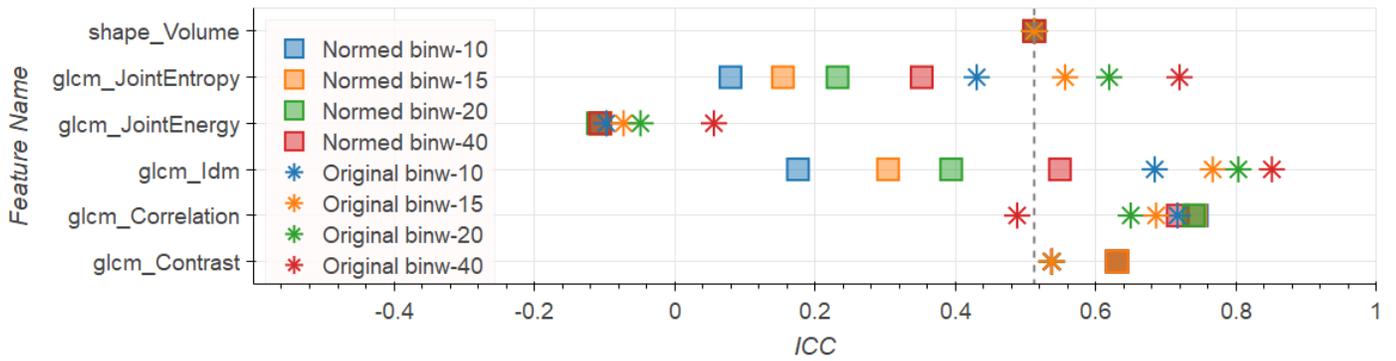
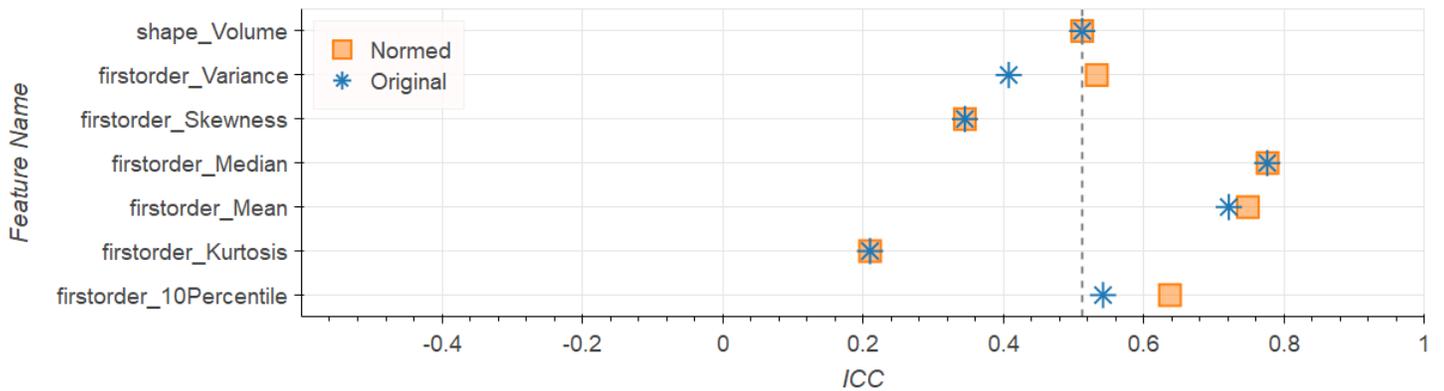

(b) Peripheral Zone (top: texture features, bottom: first order features)

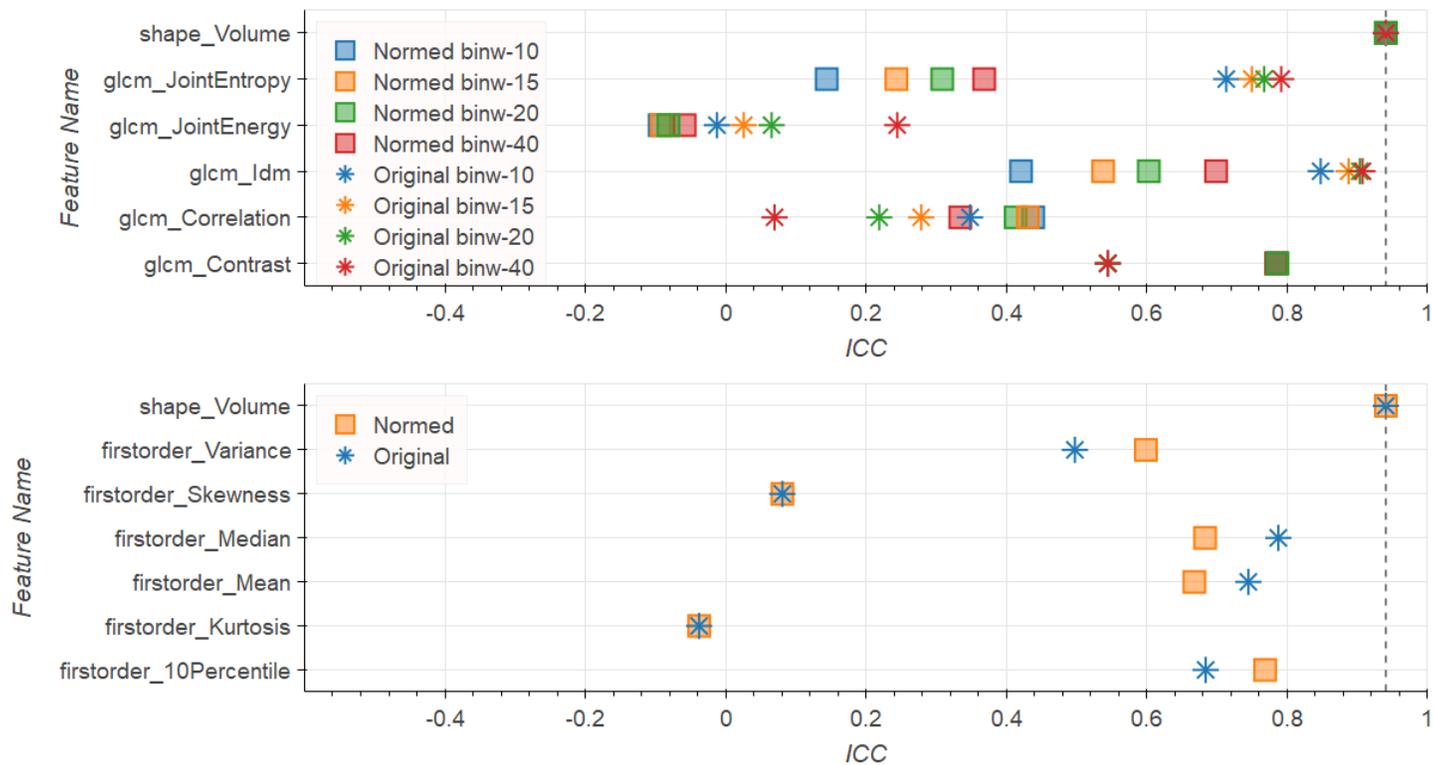

(c) Whole Gland

**Figure 7**: ICC for selected features computed on different ROIs on SUB images. Texture features are computed in 2D. Colors represent the bin width for the texture computations, glyph shape represents if the image was normalized or not. No filtering was applied to image. Dashed line indicates the reference *Volume* ICC. Results show no clear trend if normalization is better. Bin width has only marginal influence.

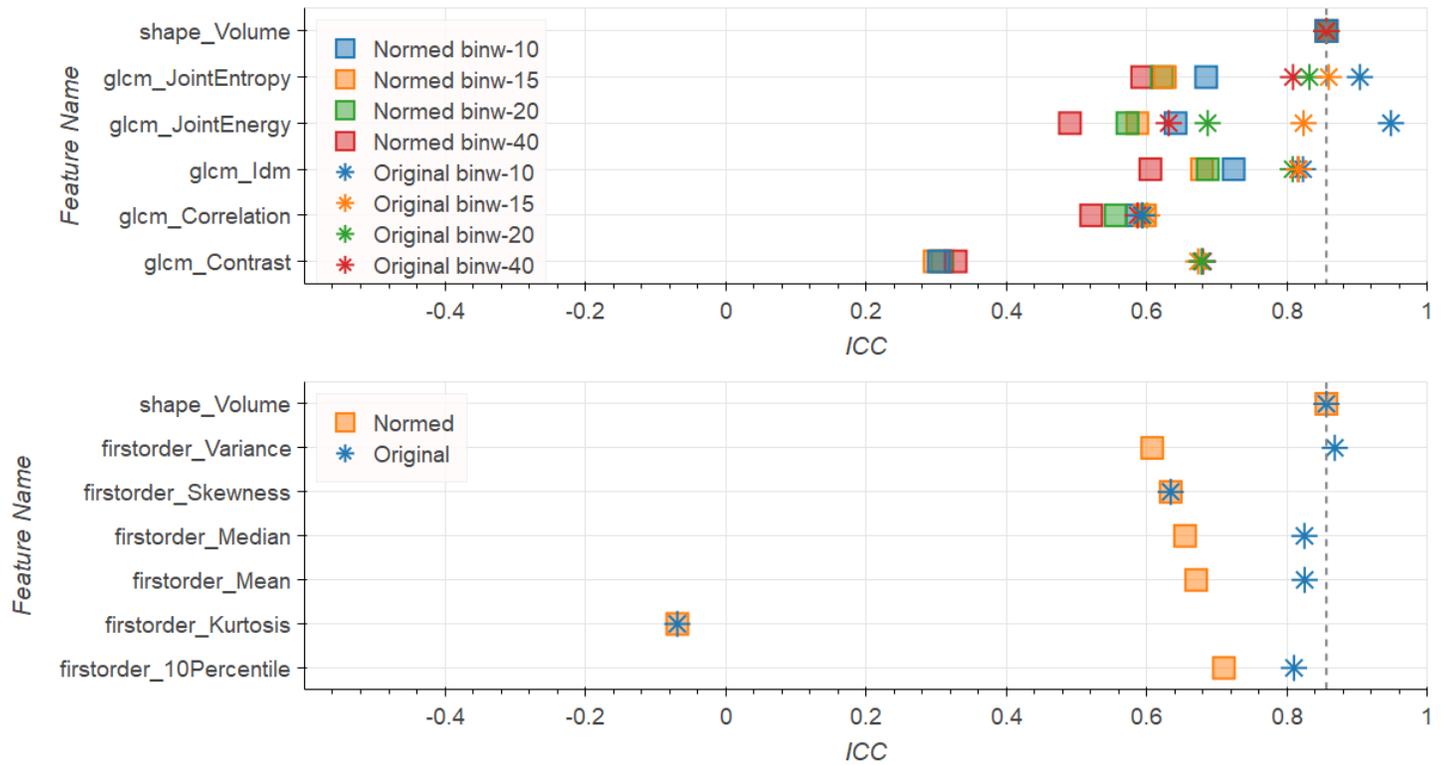

(a) Tumor ROI (top: texture features, bottom: first order features)

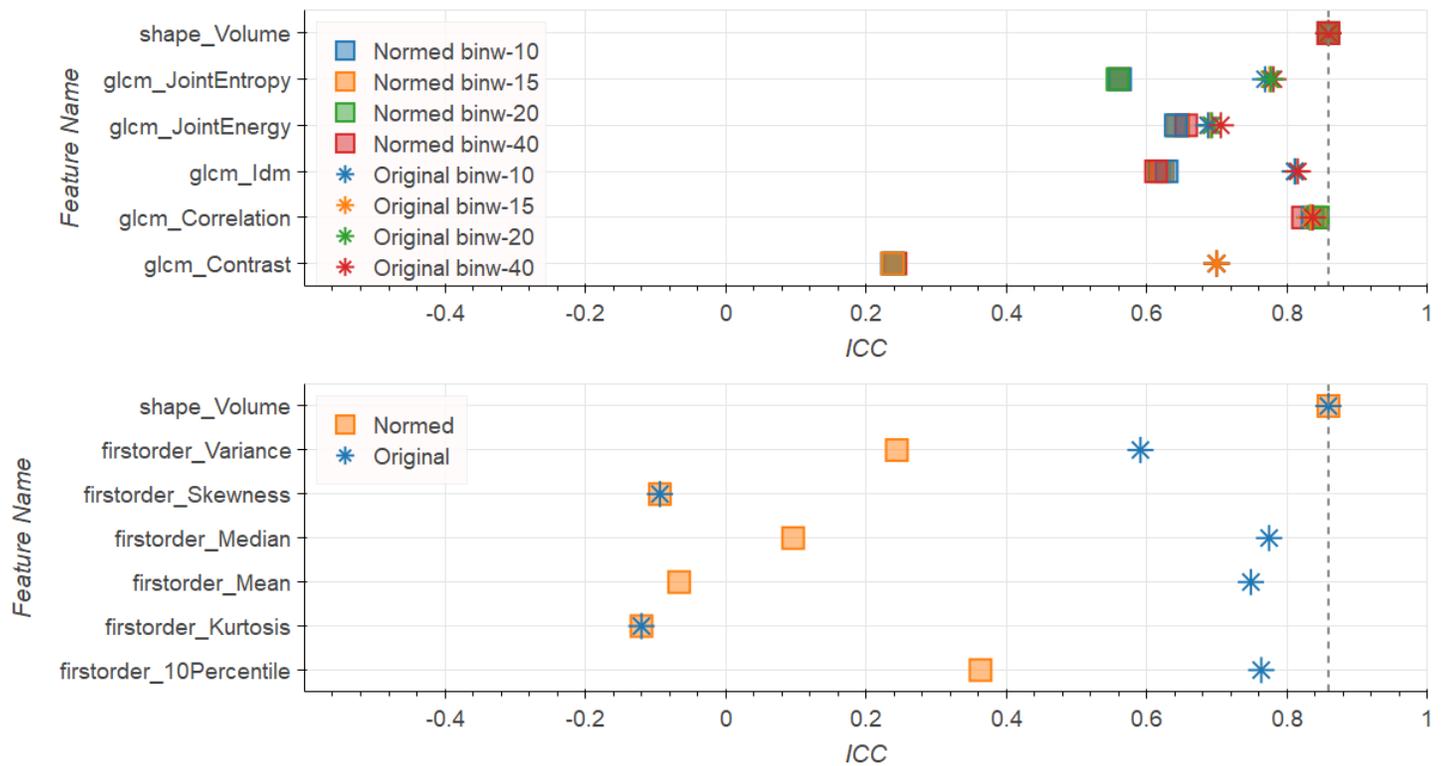

(b) Peripheral Zone (top: texture features, bottom: first order features)

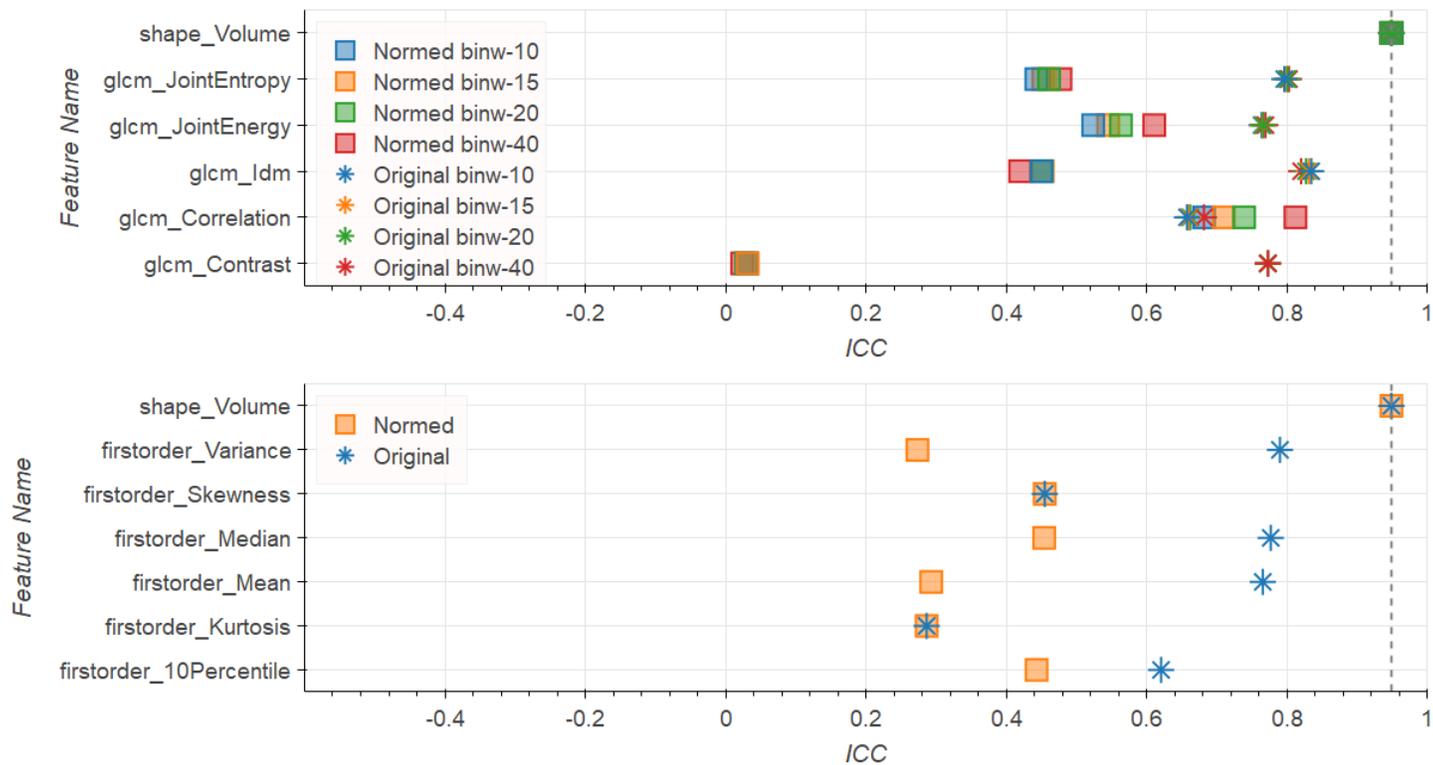

(c) Whole Gland (top: texture features, bottom: first order features)

**Figure 8**: ICC for selected features computed on different ROIs on T2w images. Texture features are computed in 2D. Colors represent the bin width for the texture computations, glyph shape represents if the image was normalized or not. No filtering was applied to image. Dashed line indicates the reference *Volume* ICC. Results show that omitting normalization rends to improve ICCs while bin width has only marginal influence.

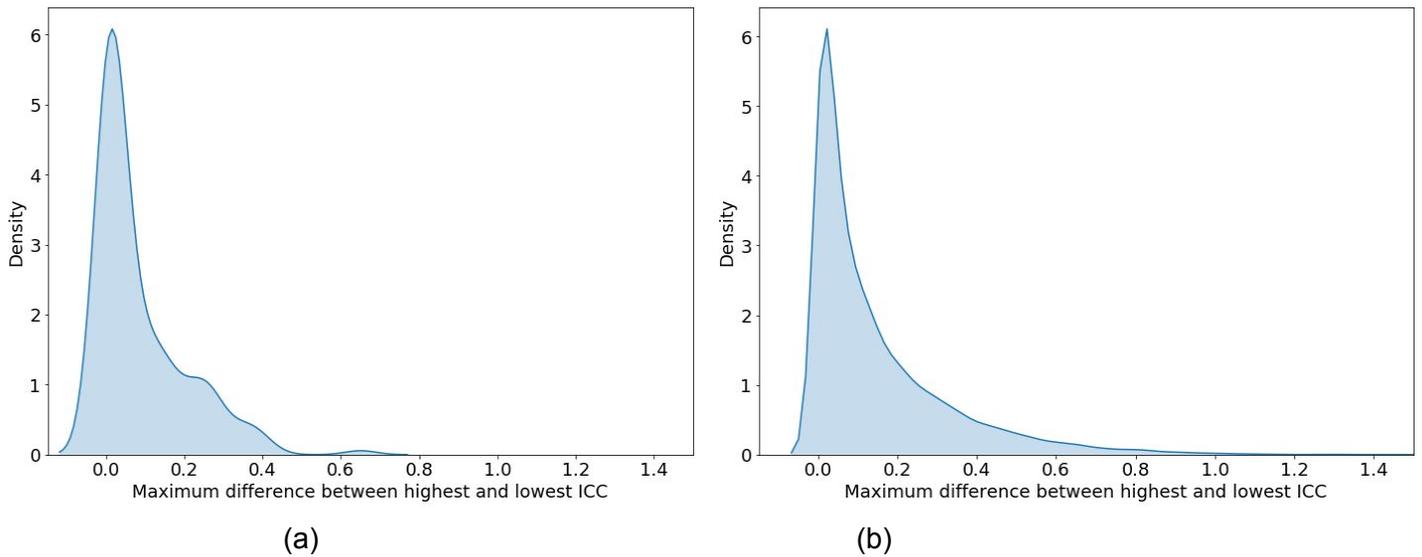

**Figure 9**: Kernel density estimation plot for the maximum difference between highest and lowest ICC per feature calculated over all considered bin widths for (a) selected GLCM texture features (Energy, Entropy, Correlation, Idm, Contrast) without pre filtering; (b) all texture features and pre filterings. Results in these figures show that for most features the maximum difference that is introduced by varying the chosen bin widths is small.

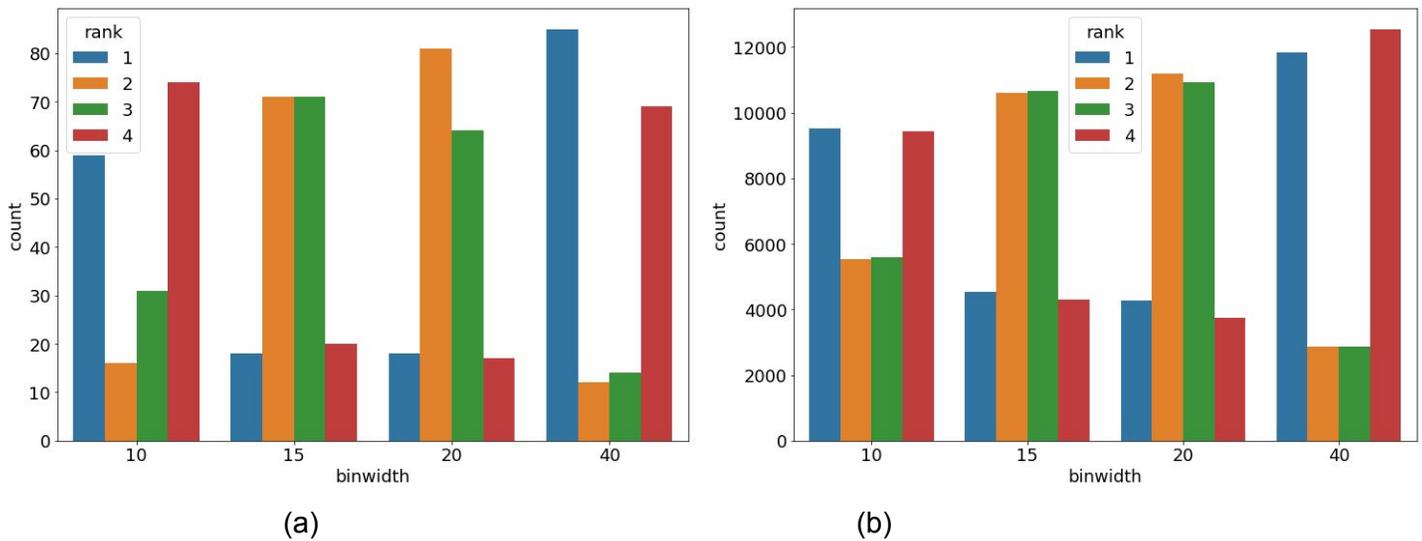

**Figure 10**: Distribution of ICC rank for texture features depending on bin width for (a) selected GLCM texture features (Energy, Entropy, Correlation, Idm, Contrast) without pre filtering; (b) all texture features and pre filterings. Results in these figures show that the lowest and highest bin width lead equally often to the best and worst ICCs among all bin widths for a feature. Hence bin widths 15 or 20 are a better choice to estimate the average repeatability of features.

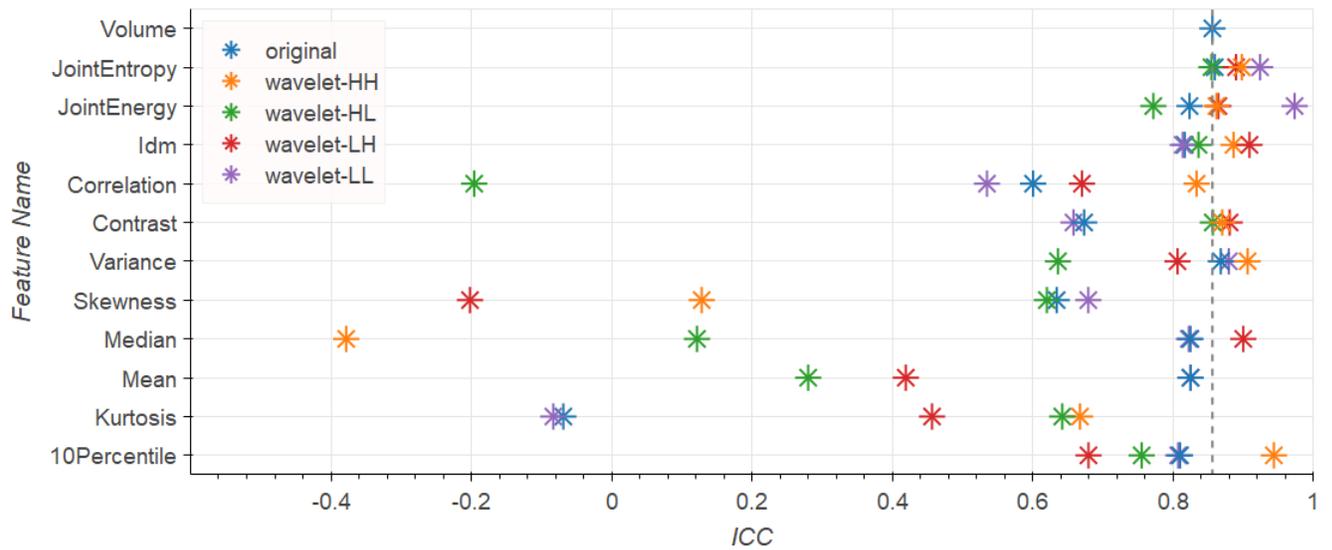

(a) 2D Wavelet Filter, 2D Texture Feature Extraction

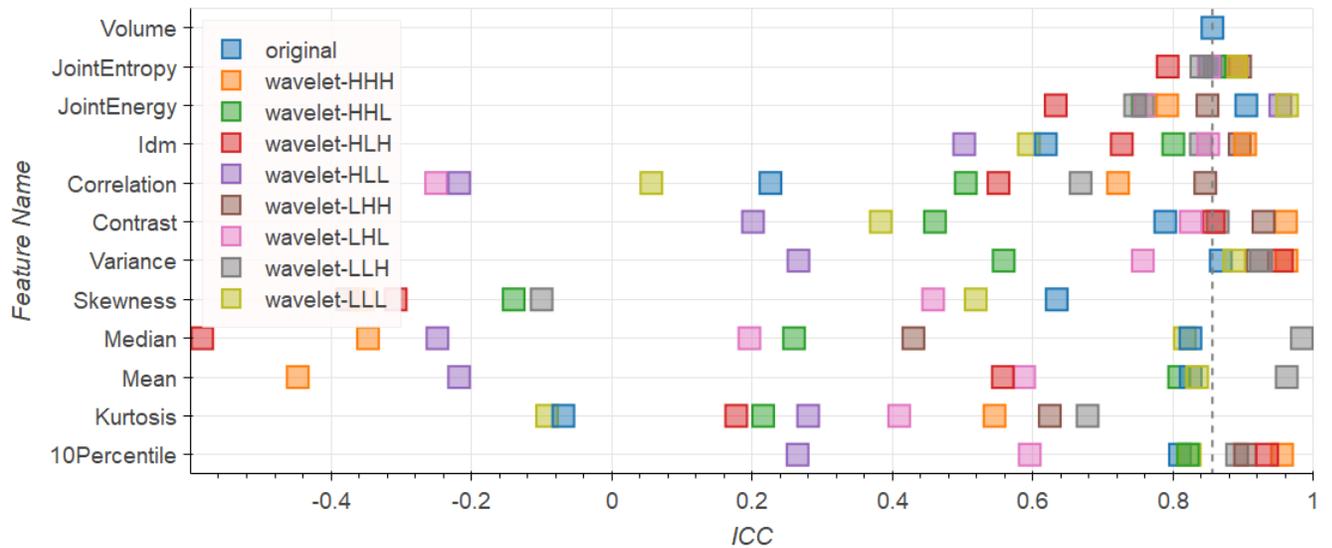

(b) 3D Wavelet Filter, 3D Texture Feature Extraction

**Figure 11**: ICC for selected features computed on the Tumor ROI on T2w images with (a) 2D and (b) 3D wavelet pre-filtering and feature extraction. Results in these figures show that some filters improve the repeatability of particular features (e.g. Contrast) to reach above-reference performance. However, no wavelet filter variant is consistently related to high ICCs across all features. Also no clear advantage of 2D over 3D (or vice versa) feature extraction can be observed.

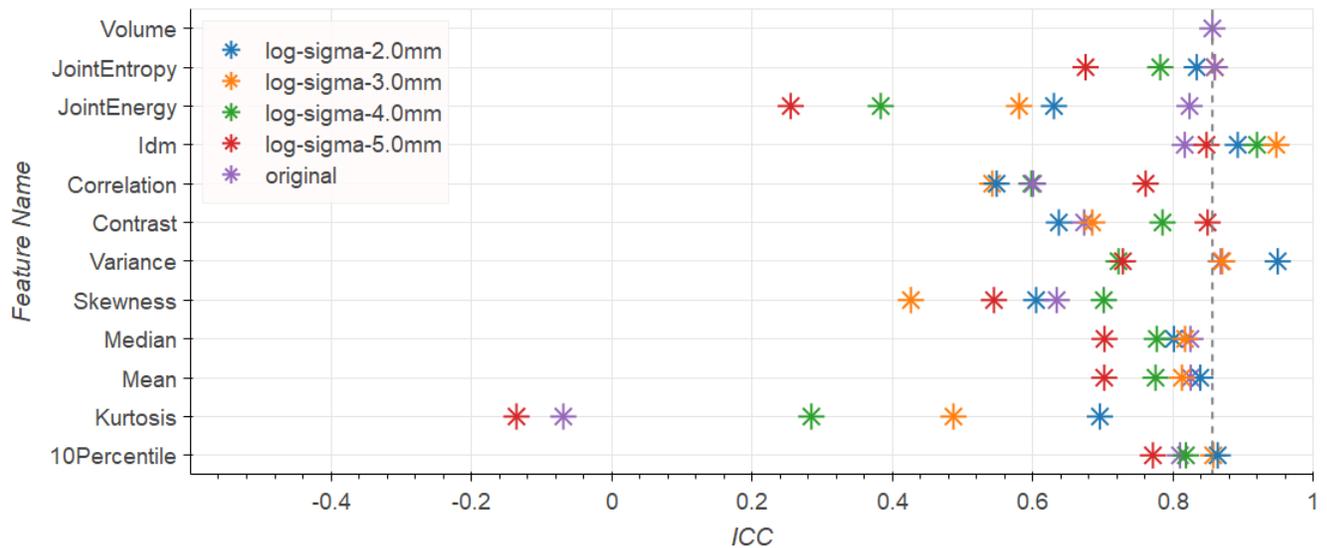

(a) 3D Laplacian of Gaussian Filters, 2D Texture Feature Extraction

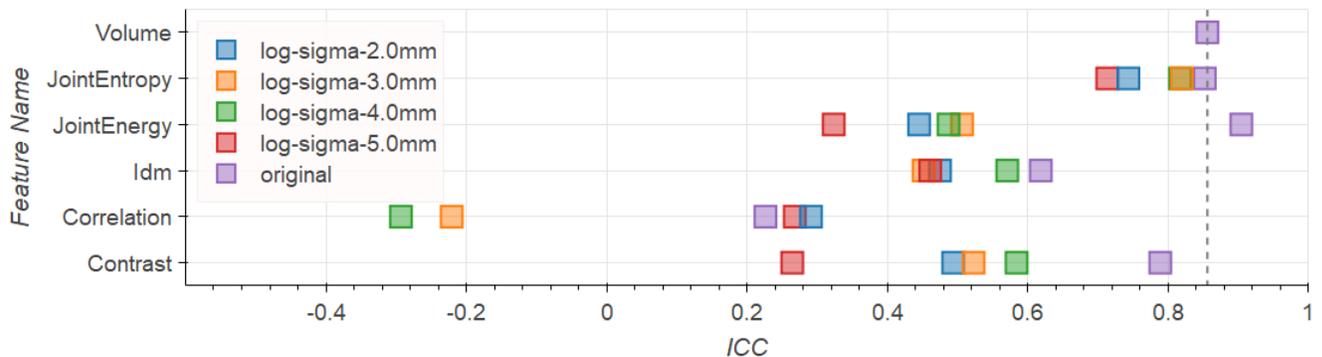

(b) 3D Laplacian of Gaussian Filters, 3D Texture Feature Extraction

**Figure 12**: ICC for selected features computed on the Tumor ROI on T2w images with 3D LoG pre-filtering and texture features extracted in 2D (a) and 3D (b). Filter legends in the plot match those produced by pyradiomics, i.e., "log-sigma-5.0mm" corresponds to LoG filter with sigma of 5.0 mm. Results in this figure do not reveal filters which consistently improve repeatability. Also no clear advantage of 2D over 3D (or vice versa) feature extraction can be observed.

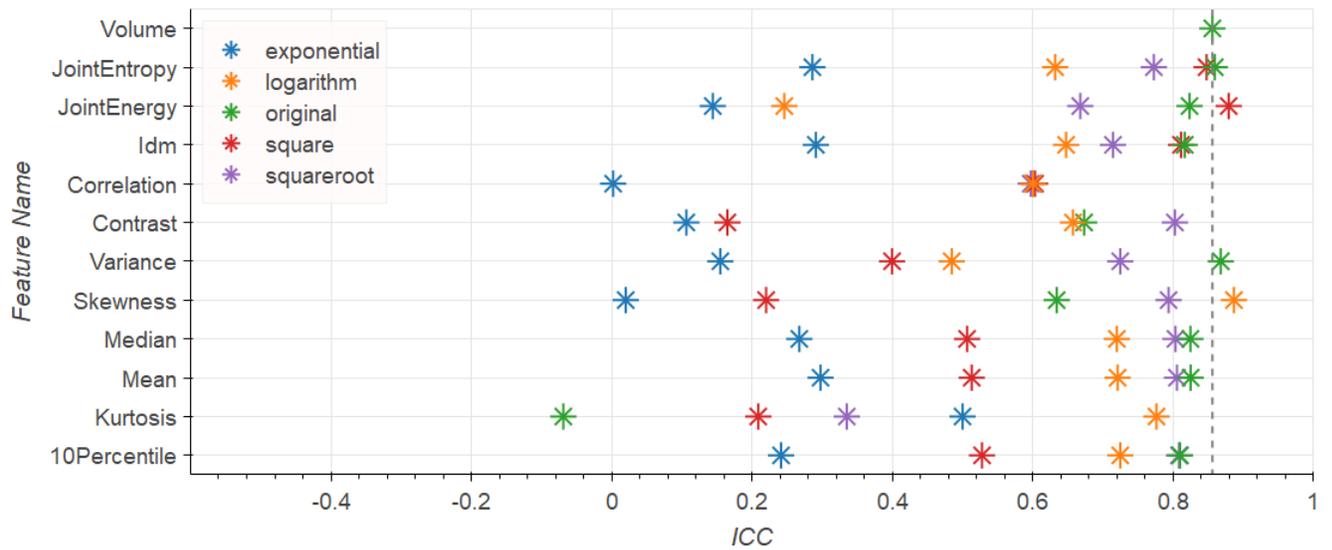

(a) Single Pixel Filters, 2D Texture Feature Extraction

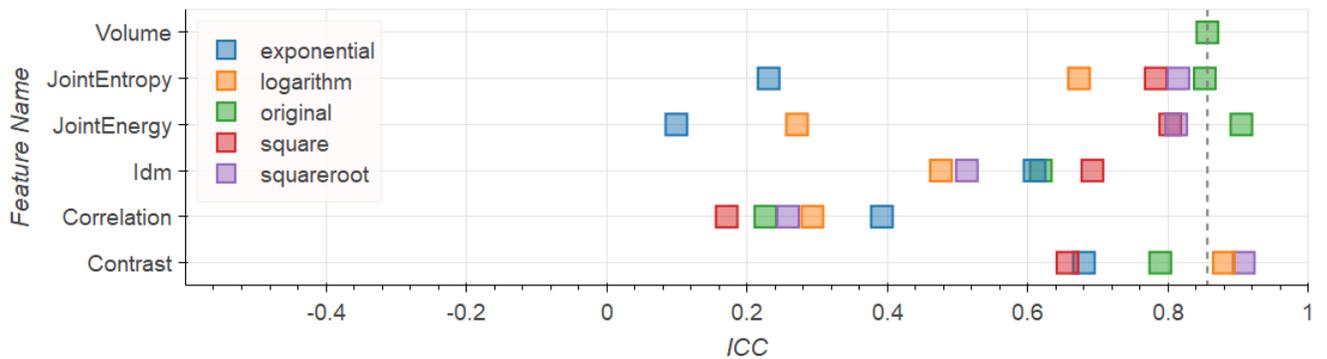

(b) Single Pixel Filters, 3D Texture Feature Extraction

**Figure 13**: ICC for selected features computed on the Tumor ROI on T2w images with single pixel pre-filtering and texture features extracted in 2D (a) and 3D (b). Results in this figure do not reveal filters which consistently improve repeatability. Also no clear advantage of 2D over 3D (or vice versa) feature extraction can be observed.

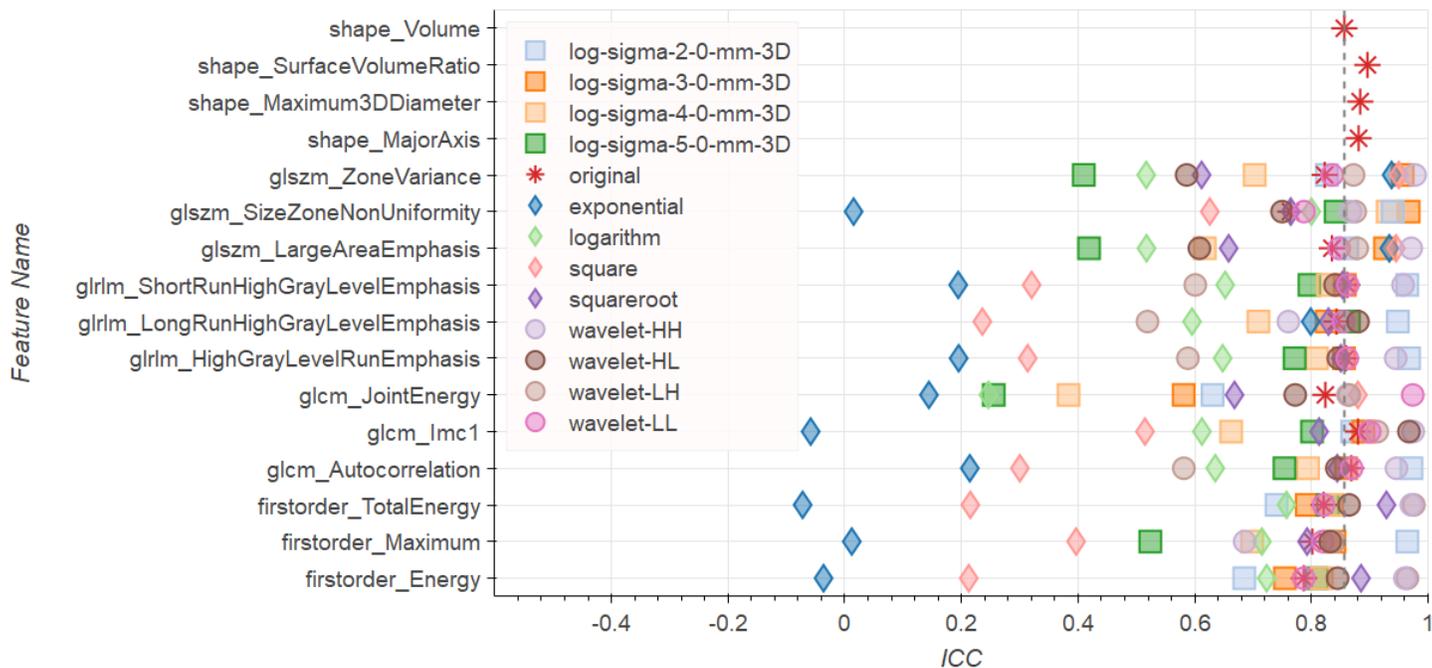

(a) T2w Tumor ROI

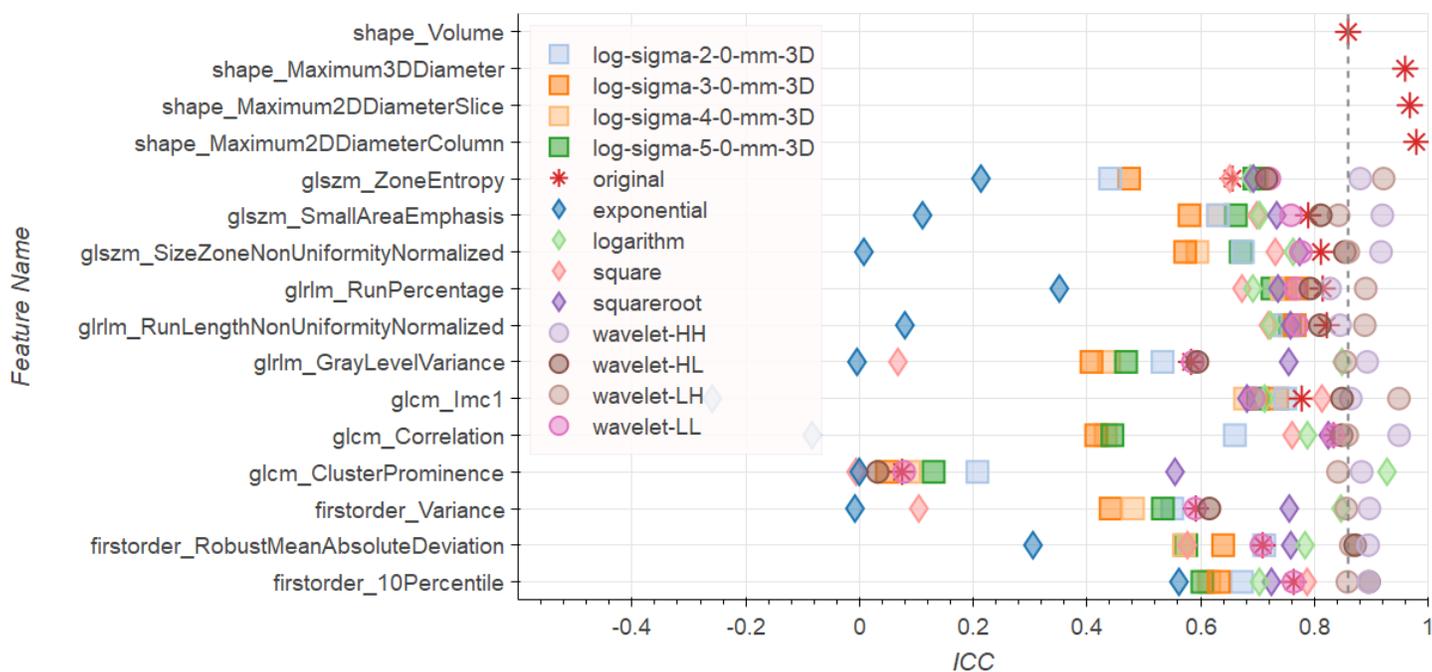

(b) T2w Peripheral Zone

**Figure 14**: Top 3 features for each feature group by ICC in T2w images for (a) the Tumor ROI and (b) the Peripheral Zone. Results in these figures illustrate that ICCs are spread over a wide range depending. Also some filters have show a consistently low performance. However, no filter consistently performs above reference. We can also see that some shape filters have a higher repeatability than *Volume*.

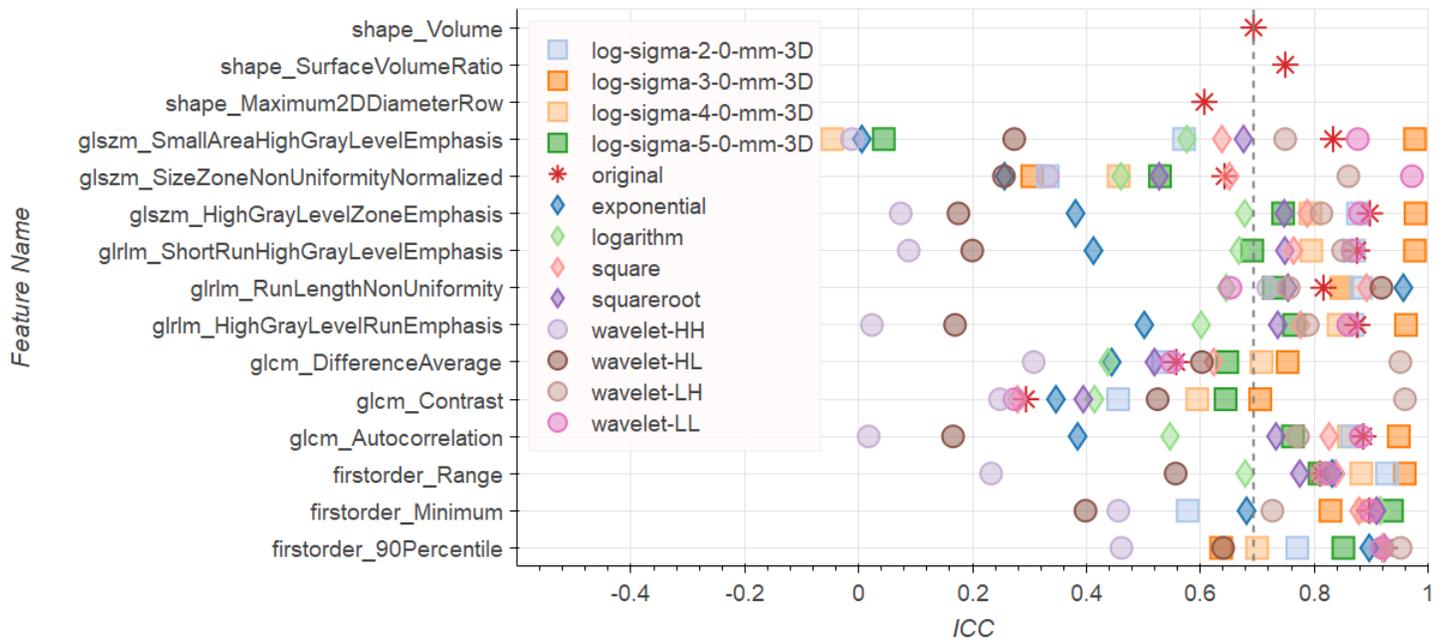

(a) ADC Tumor ROI

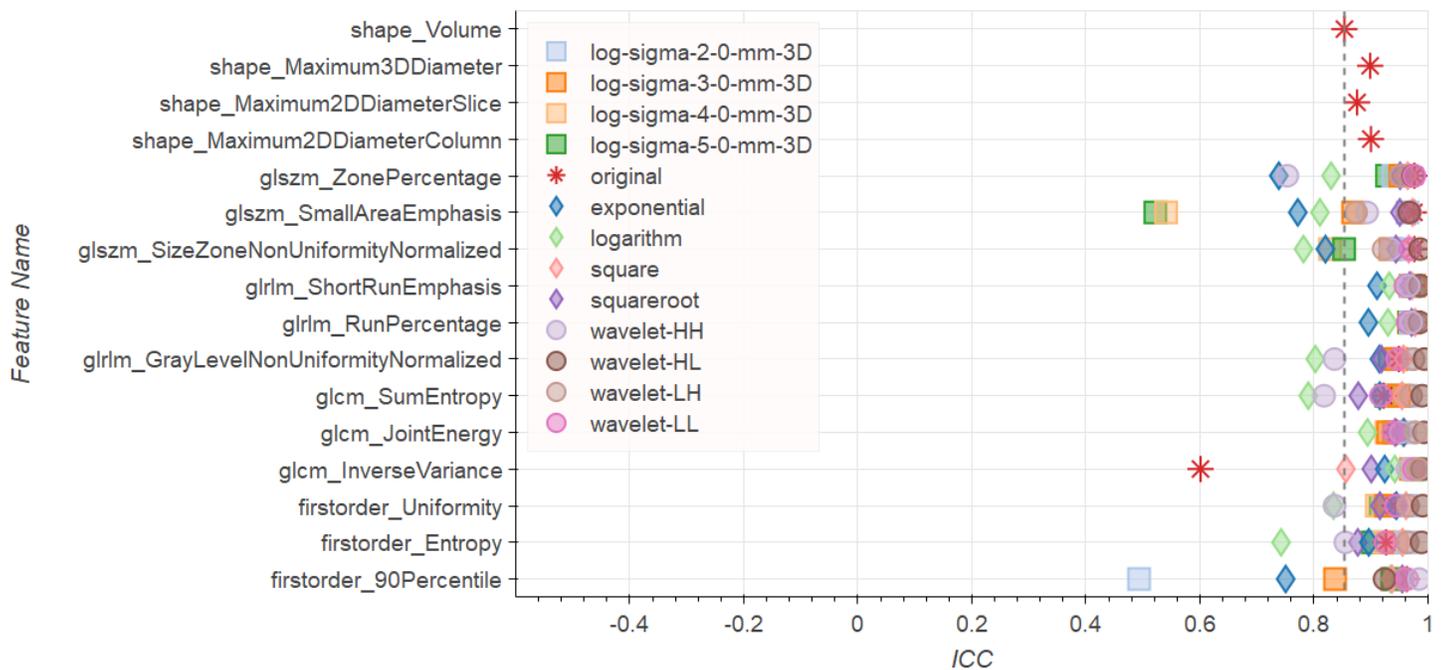

(b) ADC Peripheral Zone

**Figure 15**: Top 3 features for each feature group by ICC in ADC images for (a) the Tumor ROI and (b) the Peripheral Zone. Results in these figures show again a wide spread of ICCs in the Tumor ROI, with some filters having tendency towards high ICCs, however, not consistently. For the Peripheral Zone many filters reach consistently above reference repeatability. Overall, ICCs are also much less spread.

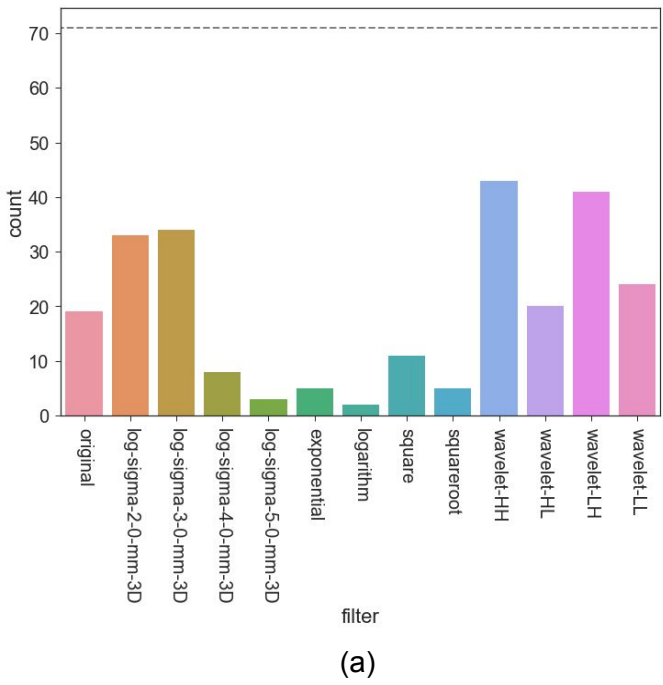
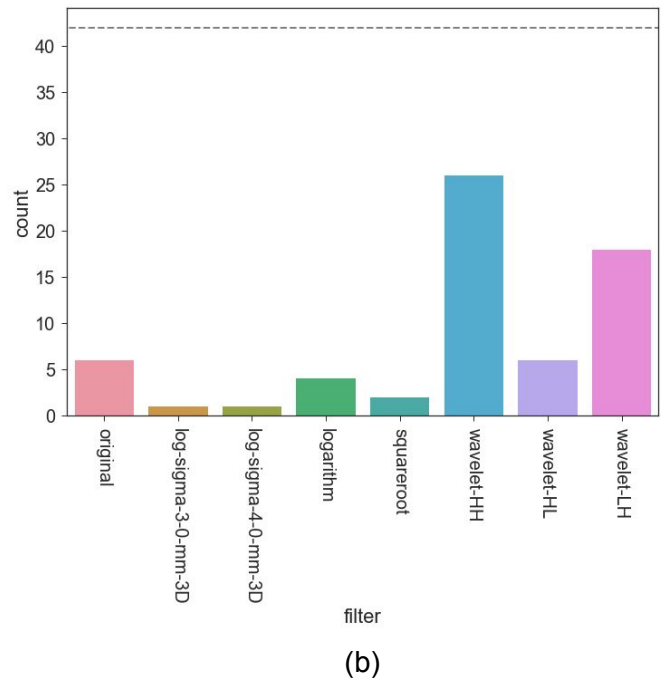

(a)                  (b)

**Figure 16**: Overview of how often the particular pre-filters appear among the features which reach an ICC higher than Volume on T2w images for (a) Tumor ROI and (b) Peripheral Zone. Dashed line indicates total number of features with an ICC higher than *Volume*. Note that for one feature several filters can appear. Results in these figures illustrate that some filters are consistently more often related to high repeatability than others. However, no filter comes even close to being always related to high repeatability.

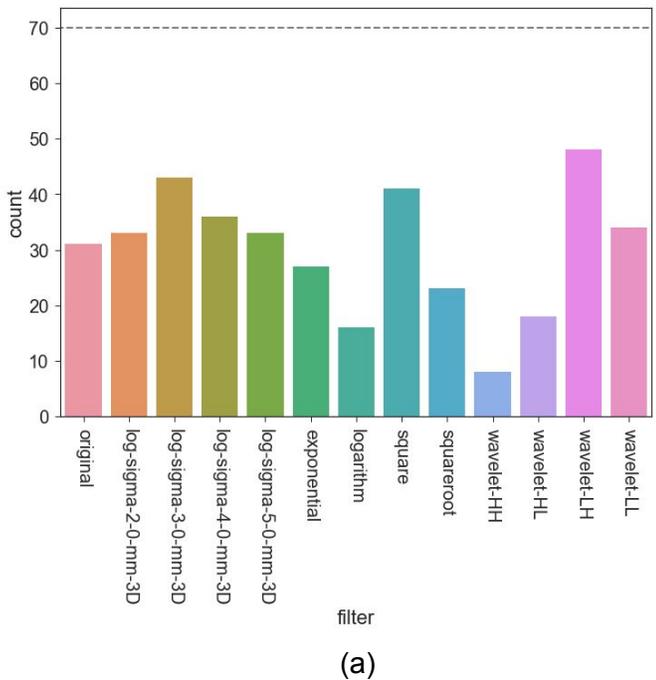
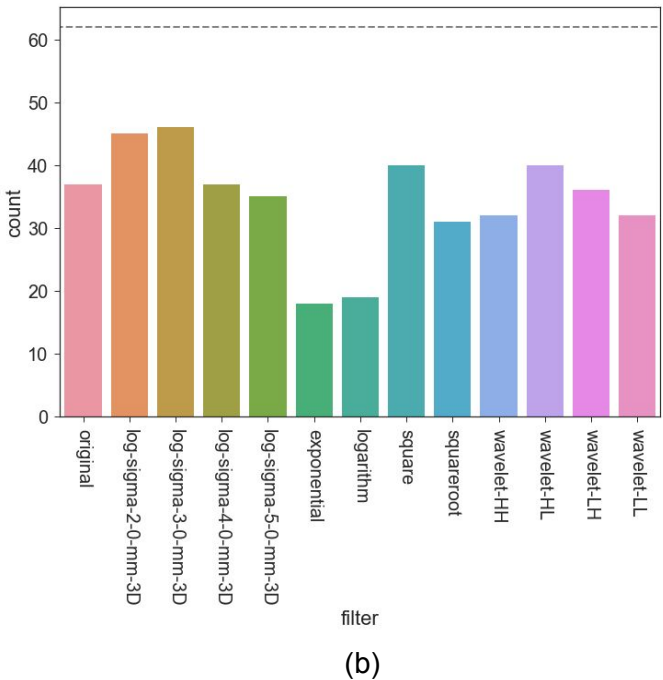

(a)                  (b)

**Figure 17**: Overview of how often the particular pre-filters appear among the features which reach an ICC higher than Volume on ADC images for (a) Tumor ROI and (b) Peripheral Zone. Dashed line indicates total number of features with an ICC higher than *Volume*. Note that for one feature several filters can appear. Results in these figures illustrate that almost all filters are related to high repeatability in about half of the above-reference group. However, no filter comes close to being always related to high repeatability.

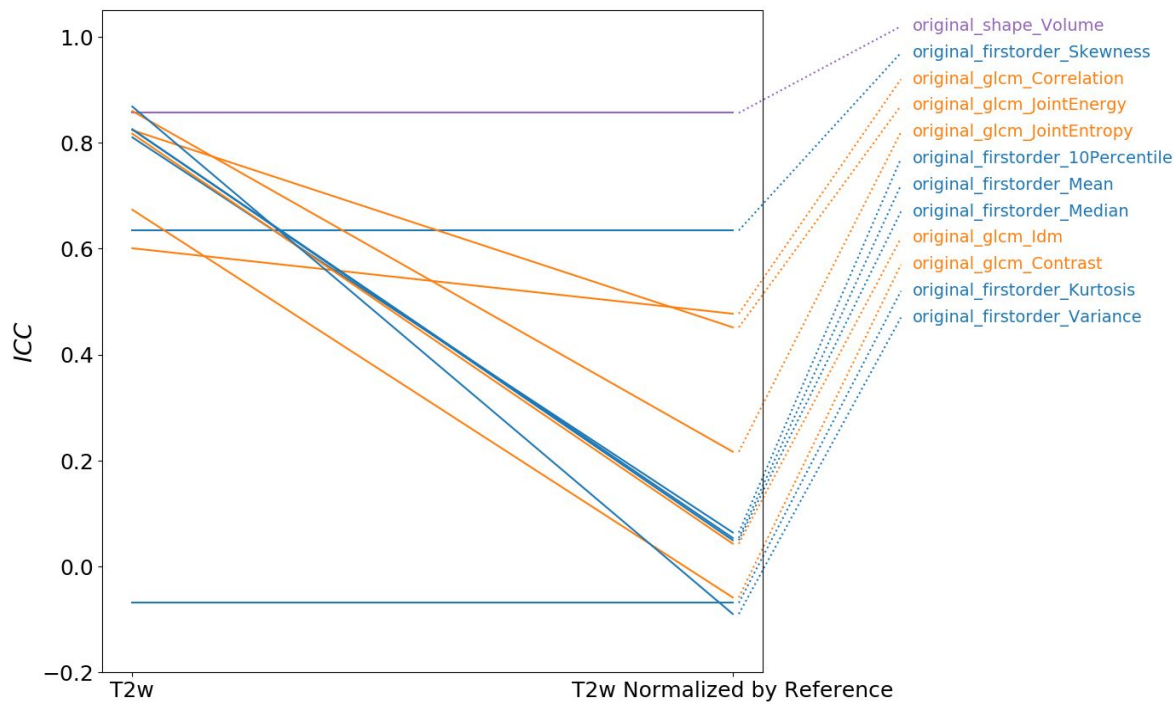

(a)

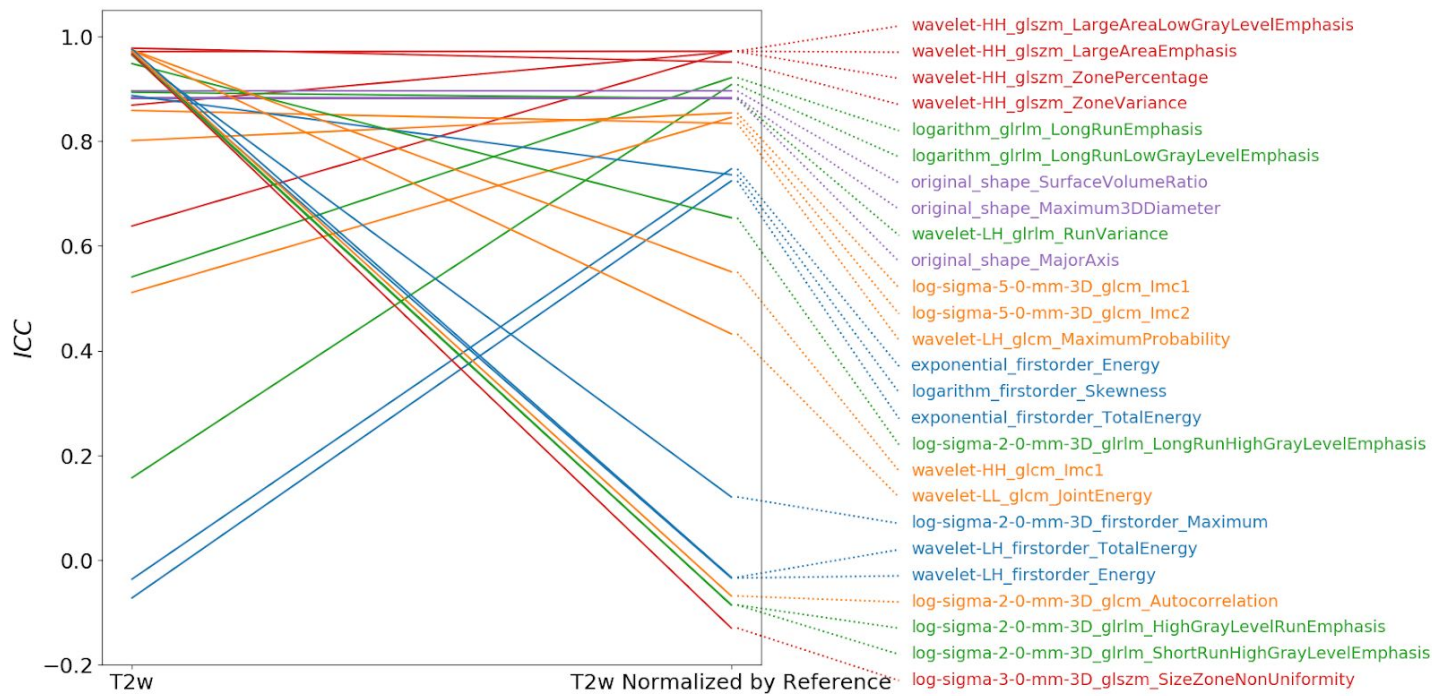

(b)

**Figure 18:** Change in Tumor ROI ICCs from T2w to T2w normalized by a reference region (muscle) for (a) literature recommended features, (b) 3 most stable features from each of the feature classes. Note that for each image configuration the top 3 features were selected (hence up to 6 features per feature group are plotted). Results in this figure show that for most features normalization by a reference region decreases repeatability.

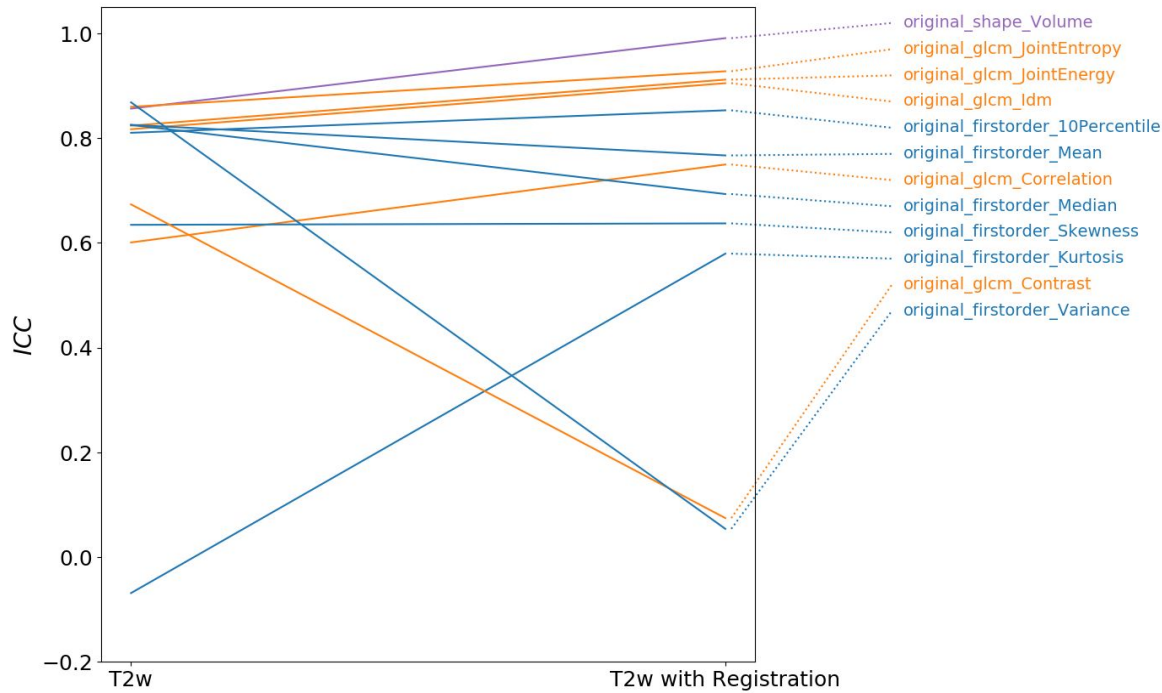

(a)

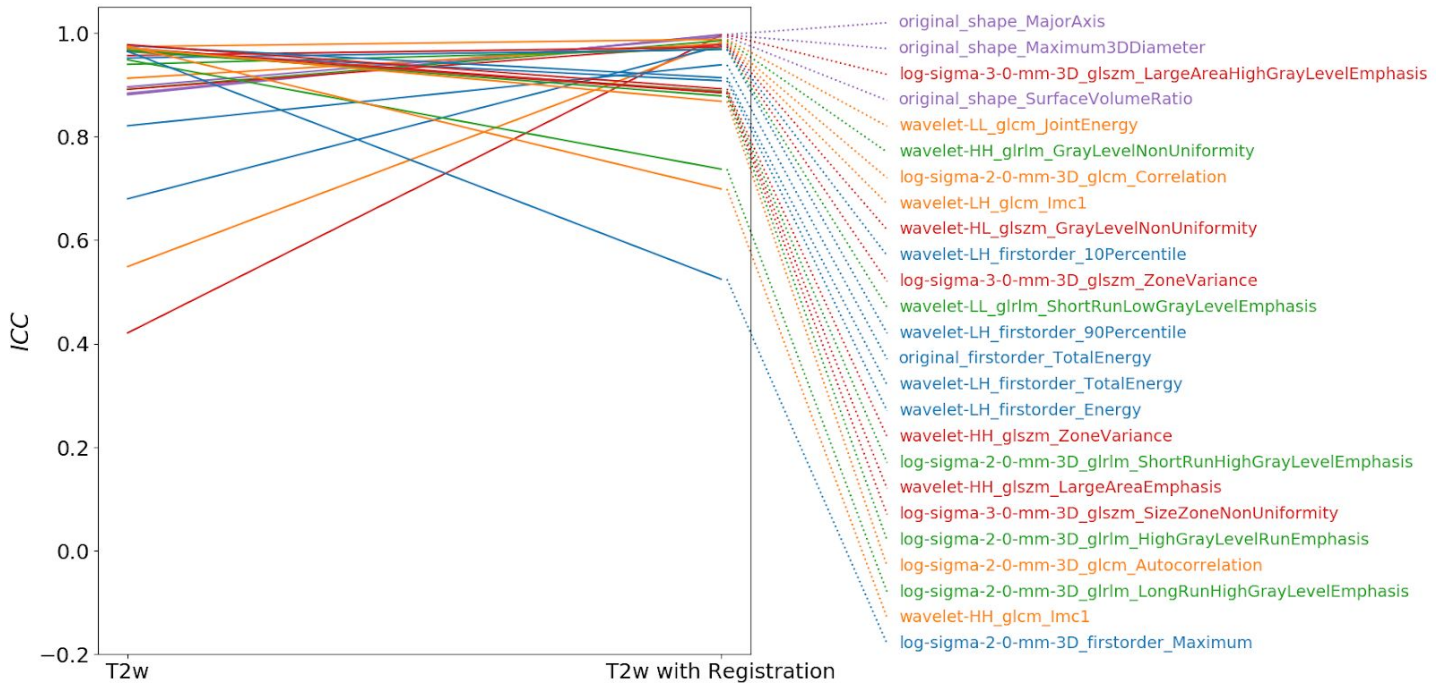

(b)

**Figure 19:** Change in ICCs from T2w Tumor ROIs to registered T2w Tumor ROIs for (a) literature recommended features, (b) top 3 features. Note that for each image configuration the top 3 features were selected (hence up to 6 features per feature group are plotted). Results in this figure show that there is no consistent improvement of repeatability by using registration as opposed to individual segmentations per time point.

# Appendix

*A - Description of Feature Data*

**File Format Description**

The files are in CSV format. Each row contains all features extracted for one image and mask combination. The organization and naming of the columns is directly taken from [pyradiomics](). We just added a few columns containing some additional meta information about the image/mask from which the features were derived.

The first few columns contain general info about the feature extraction (prefixed with "general_info"):

| Column Name (w/o prefix) | Meaning |
| --- | --- |
| BoundingBox | The bounding box considered around the mask |
| EnabledImageTypes | Indicates which pre-filtering options were activated for the extraction |
| GeneralSettings | Indicates which general settings were activated for the extraction (e.g. normalization, resampling, ...) |
| ImageHash | Unique identifier of the image |
| ImageSpacing | Voxel spacing |
| MaskHash | Unique identifier of the mask |
| NumpyVersion | Numpy version used by pyradiomics |
| PyWaveletVersion | PyWavelet version used by pyradiomics |
| SimpleITKVersion | SimpleITK version used by pyradiomics |
| Version | Pyradiomics version used for the extraction |
| VolumeNum | Number of zones (connected components) within the mask for the specified label |
| VoxelNum | Number of voxels in the mask |

Afterwards follow the columns for each feature. The feature column names follow this pattern:

*[pre-filter]_[feature group]_[feature name]*

For example:

- original_shape_Volume

- wavelet-HH_glcm_JointEnergy

At the end we added a few columns with additional meta information:

| Column Name | Meaning |
| --- | --- |
| study | Identifying the study this image belongs to |
| series | Identifying the series this image belongs to |
| canonicalType | Type/modality of the image (e.g. ADC, T2w, SUB) |
| segmentedStructure | Type of structure segmented by the mask |

**Filename Pattern Description**

The filenames of the feature data CSVs also contain some additional meta information about their content. The jupyter notebook for generating figures will parse this information and save it with the statistics it creates from the feature data (so you don't really have to worry about these to much).

The following table explains the different "codes" in the filename of a feature CSV file:

| File name contains | Meaning |
| --- | --- |
| FullStudySettings | Simply indicates that the extraction settings were according to this study (we also did smaller studies) |
| noNormalization | Indicates that the default pyradiomics whole-image normalization was deactivated |
| 2d/3D | Indicates if texture features were computed in 2D or 3D |
| biasCorrected | Indicates that we applied bias correction to the T2w images before processing them with pyradiomics |
| TP2Registered | Indicates that for the T2w images we didn't use the first manual segmentation but used registration to transfer the second timepoint masks to the first timepoint (see paper for more info) |
| MuscleRefNorm | Indicates that we normalized the T2w images against a consistent reference region in muscle tissue |
| T2AX | Contains only results for T2w images |
| bin10/bin15/bin20/bin40 | Bin size used for texture feature computation |